\documentclass[11pt]{article}

\usepackage[preprint]{acl}

\usepackage{times}
\usepackage{latexsym}

\usepackage[T1]{fontenc}

\usepackage[utf8]{inputenc}

\usepackage{microtype}

\usepackage{inconsolata}

\usepackage{graphicx}

\usepackage{enumitem}
\usepackage{booktabs}
\usepackage{amsmath,amssymb}
\usepackage{algorithm}
\usepackage{algpseudocode}
\usepackage{hyperref}
\usepackage{cleveref}
\usepackage{mathrsfs}
\usepackage{wrapfig}
\usepackage{subfigure}
\usepackage{subcaption}
\usepackage{xcolor}
\usepackage{colortbl} 
\usepackage{titletoc}

\usepackage[most]{tcolorbox}
\usepackage{lipsum}
\usepackage{listings}
\usepackage{caption}

\definecolor{headergray}{RGB}{60, 60, 60}       
\definecolor{baselineblue}{RGB}{230, 240, 250}  
\definecolor{ourgreen}{RGB}{235, 250, 235}      
\definecolor{codebg}{gray}{0.95}                
\definecolor{framegray}{gray}{0.6}              

\tcbset{
    enhanced,
    colback=white,
    boxrule=0.5pt,
    colframe=framegray,
    fonttitle=\bfseries\small\sffamily,
    arc=2mm,
    drop shadow=black!30!white, 
    left=6pt, right=6pt, top=6pt, bottom=6pt
}

\title{Unlocking Exploration in RLVR: Uncertainty-aware Advantage Shaping for Deeper Reasoning}

\author{\textbf{Can Xie, Ruotong Pan, Xiangyu Wu, Yunfei Zhang,}\\
\textbf{Jiayi Fu, Tingting Gao, Guorui Zhou}\thanks{Corresponding author}\\[5pt]
Kuaishou Technology\\[3pt]
 \texttt{\shortstack[c]{%
  xiecan02@gmail.com\\
  \{panruotong03, wuxiangyu06, zhangyunfei05, fujiayi,\\
  lisize, zhouguorui\}@kuaishou.com}}
}

\begin{document}
\maketitle
\begin{abstract}
Reinforcement Learning with Verifiable Rewards (RLVR) has shown significant promise for enhancing the reasoning capabilities of large language models (LLMs). However, prevailing algorithms like GRPO broadcast a uniform advantage signal across all tokens in a sequence. This coarse-grained approach overlooks the pivotal role of uncertain, high-stakes decisions during reasoning, leading to inefficient exploration and the well-documented problem of entropy collapse. To address this, we introduce \textbf{U}n\textbf{C}ertainty-aware \textbf{A}dvantage \textbf{S}haping (\textbf{UCAS}), a model-free method that refines credit assignment by leveraging the model's internal uncertainty signals. UCAS operates in two stages: it first modulates the response-level advantage using a logit-space self-confidence proxy, and then applies an asymmetric token-level penalty based on raw logit certainty. This dual mechanism encourages exploration of high-uncertainty paths that yield correct answers while penalizing overconfident yet erroneous reasoning, effectively balancing the exploration-exploitation trade-off. Extensive experiments on five mathematical reasoning benchmarks show that UCAS significantly outperforms strong RLVR baselines across multiple model scales, including 1.5B and 7B. Our analysis confirms that UCAS not only achieves higher rewards but also promotes greater reasoning diversity and successfully mitigates entropy collapse. Code is available at \url{https://github.com/xvolcano02/UCAS}.
\end{abstract}

\section{Introduction}

Reinforcement learning (RL) has recently become a cornerstone for enhancing the complex reasoning abilities of Large Language Models (LLMs), moving beyond simple pattern matching toward more robust problem-solving. Among the various RL approaches, Reinforcement Learning with Verifiable Rewards (RLVR) has proven particularly effective. In this paradigm, a policy model explores a vast solution space and receives feedback from verifiable signals, such as the correctness of a final answer in mathematical reasoning. This direct feedback loop has enabled policy optimization algorithms like Group Relative Policy Optimization (GRPO) \citep{shao2024deepseekmath} to achieve substantial performance gains, powering state-of-the-art systems such as DeepSeek-R1 \citep{guo2025deepseek}.

However, the success of RLVR reveals a critical underlying tension: the trade-off between precision and diversity. While methods like GRPO excel at increasing the probability of generating correct answers, they often do so at the cost of exploration. Due to the absence of a critic model, the learning signal in GRPO, which applies a single uniform advantage across all tokens, provides an indiscriminate and overly coarse form of credit assignment. It rewards all steps of a correct path equally and penalizes all steps of an incorrect one, failing to distinguish crucial reasoning leaps from trivial ones. This coarse-grained feedback drives the policy to converge prematurely on a small set of "safe" high-reward trajectories. A common side effect is entropy collapse \citep{cui2025entropy}, where the output distribution contracts, reducing solution diversity and impairing performance on complex problems that demand novel reasoning strategies.

Previous studies \citep{wang2023math, lightman2024lets, chen2024step, zhang2024generative, sun2025efficient} have attempted to employ process-level reward models to deliver more fine-grained signals. However, as DeepSeek \citep{guo2025deepseek} points out, training fine-grained reward models is costly, difficult to scale, limited in its ability to provide accurate signals, and vulnerable to reward hacking. Some recent efforts \citep{chen2025seed, cheng2025reasoning, wang2025beyond} have tried to incorporate entropy-based feedback to enhance advantages, such as integrating semantic entropy or policy entropy related to the response into advantage calculations. Yet, most studies either pursue low entropy to improve accuracy or encourage high entropy to maintain exploration, lacking fine-grained modeling of the relationship between responses and their policy entropy.

To address the above challenges, we propose an \textbf{U}n\textbf{C}ertainty-aware \textbf{A}dvantage \textbf{S}haping (\textbf{UCAS}), a model-free method that refines credit assignment in RLVR by leveraging the model’s intrinsic uncertainty. UCAS is designed to resolve the precision–diversity dilemma by reshaping the advantage signal at two complementary levels. At the response level, UCAS modulates the sequence-level advantage using a logit-space self-confidence proxy, amplifying rewards for correct-but-uncertain responses and penalties for incorrect-but-confident ones. At the token level, it further introduces an asymmetric certainty-based penalty derived directly from raw logits, suppressing overconfident errors while safeguarding positive reasoning trajectories from negative token-level updates. Collectively, these mechanisms promote exploration of uncertain but potentially fruitful reasoning paths, while efficiently suppressing confidently wrong solutions. Extensive experiments on five mathematical reasoning benchmarks demonstrate that UCAS consistently outperforms strong RLVR baselines at both the 1.5B and 7B model scales. Beyond reward improvements, UCAS fosters greater reasoning diversity and substantially mitigates entropy collapse, confirming the effectiveness of uncertainty as a fine-grained learning signal.

Our contributions can be summarized as follows:
\begin{itemize}
\item We propose UCAS, an extra-model-free fine-grained advantage shaping mechanism based on internal confidence signals, which performs uncertainty-aware advantage adjustment at both response and token levels.
\item We provide a novel mechanism to adaptively calibrate advantages based on uncertainty, enabling steady reward gains, longer reasoning chains, and entropy recovery, thus preventing entropy collapse in RLVR and improving reasoning accuracy.
\item Extensive experiments on multiple mathematical reasoning benchmarks demonstrate that UCAS significantly improves model reasoning performance, validating its effectiveness in enhancing exploration diversity and optimization outcomes.
\end{itemize}

\section{Background: Reinforcement Learning with Verifiable Rewards}

In the training of large language models, early mainstream reinforcement learning alignment methods primarily relied on PPO. By introducing a clipping ratio into the objective function, PPO stabilizes training by constraining the magnitude of policy updates. This method has been widely adopted in Reinforcement Learning from Human Feedback (RLHF), where reward models provide preference-based scores to gradually shape model behavior. However, PPO exhibits key limitations: it depends on critic-based value estimation and requires large-scale preference annotation, both of which are costly and prone to noise accumulation.

To overcome these limitations, recent research has introduced RLVR.
RLVR converts open-ended outputs into programmatically checkable signals, such as numerical consistency in mathematics, unit-test pass rates in code generation, or formal constraint satisfaction \citep{su2025crossing,wang2025reinforcement,lin2026cec,li2026mol,jiang2026foeforesterrorsmakes}, thereby avoiding the noise and cost of preference models. By forming a closed loop of model–environment–verifier, RLVR enables policies to be updated directly from binary or graded correctness signals, improving both sample efficiency and reproducibility in structured reasoning tasks.

In the concrete implementation of RLVR, GRPO \citep{shao2024deepseekmath} emerges as a representative algorithm. Unlike PPO, which relies on critic-based value estimation, GRPO computes advantages by normalizing group-level verifiable rewards and updates the policy directly. 

Formally, the objective is given by:
\begin{equation}
\begin{aligned}
\mathcal{J}_{\text{GRPO}}(\theta) 
&= \mathbb{E}_{q \sim \mathcal{D}, \; o \sim \pi_{\theta_{\text{old}}}} \Bigg[ \frac{1}{G} \sum_{i=1}^G \frac{1}{|o_i|} \sum_{t=1}^{|o_i|} \\
& \quad \min\Big(r_{i,t}(\theta)\hat{A}_{i,t}, \\
& \quad \text{clip}(r_{i,t}(\theta),1-\epsilon,1+\epsilon)\hat{A}_{i,t}\Big) \\
& \quad - \beta D_{\text{KL}}(\pi_\theta \| \pi_{\text{ref}}) \Bigg]
\end{aligned}
\end{equation}

where 
\begin{equation}
r_{i,t}(\theta) = \frac{\pi_\theta(o_{i,t} \,|\, q, o_{i,<t})}{\pi_{\theta_{\text{old}}}(o_{i,t} \,|\, q, o_{i,<t})},
\end{equation}
denotes the probability ratio between the new and old policies for token $o_{i,t}$, and the advantage $\hat{A}_{i,t}$ is estimated from group rewards as:
\begin{equation}
\hat{A}_{i,t} = \frac{R_i - \mu(R)}{\sigma(R) + \epsilon},
\end{equation}
with $R_i$ the cumulative verifiable reward of trajectory $o_i$, $\mu(R)$ and $\sigma(R)$ the mean and standard deviation across the sampled group, and $\epsilon$ a small constant for numerical stability.

By eliminating dependency on value models and instead exploiting group-normalized verifiable rewards, GRPO achieves stable and cost-efficient training.

Building on GRPO, Decouple Clip and Dynamic Sampling Policy Optimization (DAPO) \citep{yu2025dapo} is proposed to further improve stability and exploration. 
DAPO integrates four key techniques: Clip-Higher, Dynamic Sampling, Token-Level Policy Gradient Loss, and Overlong Reward Shaping. 
Similar to GRPO, DAPO samples multiple responses per prompt and optimizes the following objective:

\begin{equation}
\begin{aligned}
&\mathcal{J}_{\text{DAPO}}(\theta) 
= \mathbb{E}_{\substack{(q,a) \sim \mathcal{D} \\ \{o_i\} \sim \pi_{\theta_{\text{old}}}}} \Bigg[ \frac{1}{\sum_{j=1}^G |o_j|} \sum_{i=1}^G \sum_{t=1}^{|o_i|} \\
& \quad \min \Big( r^i_t(\theta)\hat{A}^i_t, \text{clip}(r^i_t(\theta), \text{clip\_range})\hat{A}^i_t \Big) \Bigg], \\
&\text{s.t. } 0 < |\{ i \mid \text{is\_equiv}(o^i, a) \}| < G
\end{aligned}
\end{equation}

where $\epsilon_{\text{low}}$ and $\epsilon_{\text{high}}$ denote the lower and upper bounds of the clipping range. 
Compared to GRPO, DAPO explicitly decouples the clipping bounds, incorporates adaptive sampling strategies, thereby alleviating entropy collapse and improving the generalizability of RLVR-trained models.

\section{Method}
\label{sec:method}
To address the coarse credit assignment problem in RLVR, we introduce \textbf{Uncertainty-aware Advantage Shaping (UCAS)}, a method designed to replace the blunt instrument of uniform advantage with a more nuanced, two-stage mechanism. The central idea is to reshape the learning signal by considering uncertainty at two distinct granularities: the entire reasoning path (response-level) and the individual generative steps within it (token-level). This hierarchical approach first sets a \textit{strategic} learning objective by evaluating the value of the overall trajectory, and then \textit{locally} refines the policy update to encourage robust exploration and prevent the premature convergence that leads to entropy collapse.

\begin{figure*}[t] 
\centering
    \includegraphics[width=0.99\textwidth]{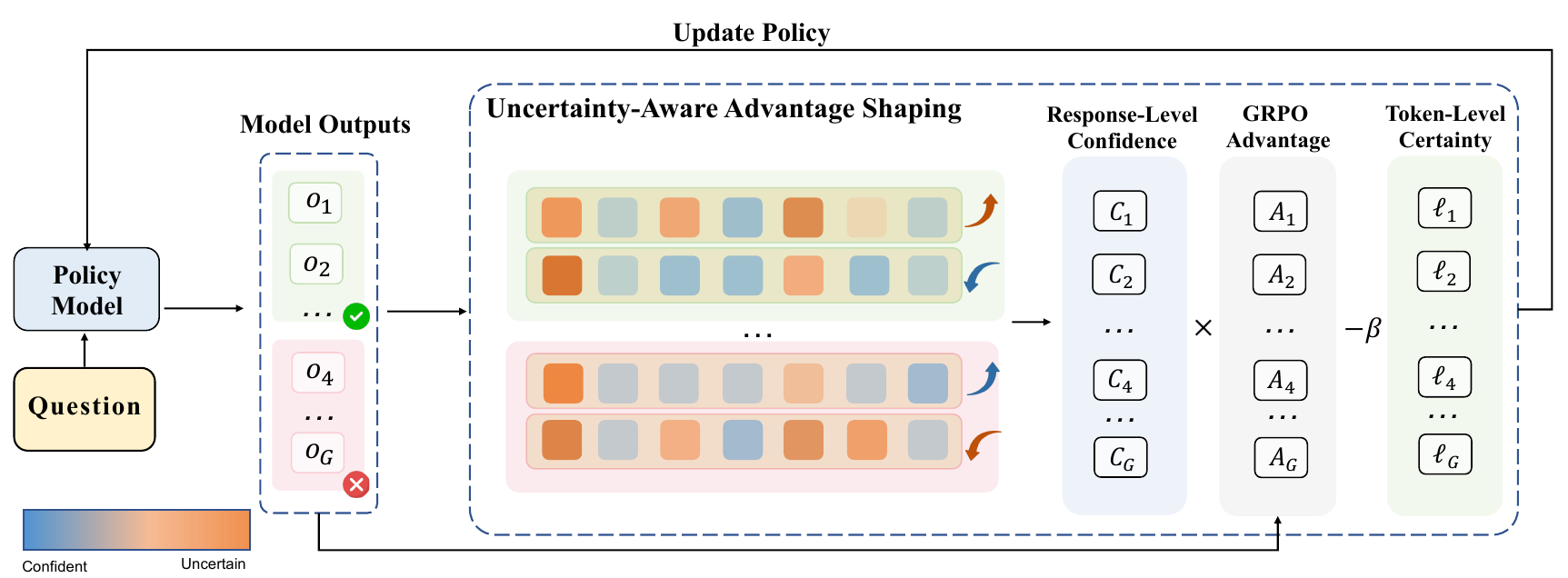}
\caption{\textbf{Overview of the UCAS Advantage Shaping Mechanism.} UCAS refines the uniform GRPO advantage through a two-stage process. \textbf{Stage 1 (Macro-level)}: It applies Response-Level Modulation using the trajectory's overall self-confidence to determine its strategic value for exploration vs. exploitation. \textbf{Stage 2 (Micro-level)}: It introduces a Token-Level Certainty Penalty using raw logits to discourage local overconfidence. The final shaped advantage $\hat{A}^{\text{UCAS}}_{i,t}$ guides a more nuanced policy update.}
\label{fig:overview}
\end{figure*}

\subsection{Uncertainty Signals: From Confidence to Logits}
\label{sec:uncertainty_quantification}
To perform this hierarchical shaping, UCAS requires signals that capture the model's epistemic state at both macro and micro levels. We extract these directly from the model's intrinsic generative process, avoiding the need for auxiliary networks.

\paragraph{Response-Level Confidence.} For a high-level assessment of a full reasoning trajectory, we use a logit-space self-confidence score. Let $\ell_{i,t,v}$ denote the pre-softmax logit of vocabulary item $v \in \mathcal{V}$ at step $t$ for response $o_i$, and define the token-level confidence as
\begin{equation}
C_{i,t} := \operatorname{LSE}(\ell_{i,t}) - \frac{1}{|\mathcal{V}|}\sum_{v \in \mathcal{V}}\ell_{i,t,v},
\end{equation}
where $\operatorname{LSE}(\ell_{i,t}) = \log \sum_{v \in \mathcal{V}} \exp(\ell_{i,t,v})$. We then define sequence-level confidence as
\begin{equation}
\mathcal{C}(o_i|q) := \frac{1}{|o_i|} \sum_{t=1}^{|o_i|} C_{i,t}.
\label{eq:self_confidence}
\end{equation}
This formulation avoids probability-space instability when token probabilities approach zero. As we show in Appendix~\ref{app:theory}, $C_{i,t}$ is equivalent to $\mathrm{KL}(U(\mathcal{V}) \parallel p_{\pi_\theta}(\cdot \mid q, o_{i,<t})) + \log |\mathcal{V}|$, so larger values of $\mathcal{C}(o_i|q)$ indicate higher overall confidence and lower uncertainty.
In contrast, when we discuss entropy collapse later in the paper, we use entropy or $\mathrm{KL}(p \parallel U)$ as a macro-level diagnostic of distributional concentration. The two directions therefore play complementary roles: $\mathrm{KL}(U \parallel p)$ serves as the training-time modulation signal, whereas $\mathrm{KL}(p \parallel U)$ characterizes the collapse outcome.

\paragraph{Token-Level Certainty.} While self-confidence is effective at the sequence level, it is derived from post-softmax probabilities, which can suffer from poor calibration \citep{liu2025uncertainty, ma2025estimating}. This can cause the model to appear equally confident in different choices, masking subtle but important variations in uncertainty. To capture a more direct and sensitive signal at the token level, we use the model's raw logit value for the chosen token $o_{i,t}$ as a proxy for certainty. Let $\ell_{i,t}$ be the logit corresponding to token $o_{i,t}$ at step $t$. A higher logit value indicates greater model certainty in its choice, prior to softmax normalization.

\subsection{UCAS: Two-Stage Advantage Shaping}
\label{sec:ucas}
Given a group of $G$ responses $\{o_1, \dots, o_G\}$ to a prompt $q$, UCAS reshapes the original GRPO advantage $\hat{A}_i$ into a fine-grained, token-specific advantage $\hat{A}^{\text{UCAS}}_{i,t}$. This process unfolds in two complementary stages.

\paragraph{Stage 1: Response-Level Advantage Modulation.}
This stage adjusts the advantage for an entire response to encourage exploration of novel correct paths and suppress confident, well-trodden incorrect paths. First, we compute the self-confidence $\mathcal{C}(o_i|q)$ for each response $o_i$ in the group. To assess confidence relative to other responses in the same group, we apply z-score normalization:
\begin{equation}
\hat{\mathcal{C}}_i = \frac{\mathcal{C}(o_i|q) - \mu_{\mathcal{C}}}{\sigma_{\mathcal{C}} + \epsilon},
\end{equation}
where $\mu_{\mathcal{C}}$ and $\sigma_{\mathcal{C}}$ are the mean and standard deviation of confidence scores across the group. This standardization makes the modulation weights comparable within each prompt group and reduces dependence on model-specific raw logit scales.

We then compute a modulation weight $W(\hat{\mathcal{C}}_i)$ based on the sign of the original advantage $\hat{A}_i$, which directly encodes the correctness of the answer. Theoretically, we select an exponential form to act as a non-linear filter. This addresses the compressed variance often found in group-normalized scores, where linear rescaling fails to sufficiently distinguish ``novel'' exploration from ``routine'' exploitation. More theoretical explanation can be found in Appendix~\ref{app:theory}.
\begin{equation}
W(\hat{\mathcal{C}}_i) = 
\begin{cases} 
    \exp(-\alpha \cdot \hat{\mathcal{C}}_i) & \text{if } \hat{A}_i > 0 \quad (\text{Correct}) \\
    \exp(\alpha \cdot \hat{\mathcal{C}}_i)  & \text{if } \hat{A}_i < 0 \quad (\text{Incorrect})
\end{cases}
\label{eq:modulation_weight}
\end{equation}
where $\alpha > 0$ is a hyperparameter controlling the shaping intensity. This formulation ensures that for correct responses, lower confidence (negative $\hat{\mathcal{C}}_i$) results in a larger weight, amplifying the reward. For incorrect responses, higher confidence (positive $\hat{\mathcal{C}}_i$) results in a larger weight, amplifying the penalty. The resulting modulated advantage is $\hat{A}^{\text{mod}}_i = W(\hat{\mathcal{C}}_i) \cdot \hat{A}_i$.

\paragraph{Stage 2: Token-Level Certainty Penalty.}
Response-level modulation sets a global learning objective for each trajectory, but this modulated advantage, $\hat{A}^{\text{mod}}_i$, is still a uniform signal broadcast to all tokens within that sequence. This alone is insufficient to prevent the model from developing localized overconfidence---a key driver of entropy collapse. The second stage therefore introduces a token-specific penalty to directly address this. The penalty is applied asymmetrically: it strongly suppresses locally overconfident tokens on unfavorable trajectories, while positive trajectories are protected from being flipped into negative token-level updates.

We use the raw logit $\ell_{i,t}$ as our certainty measure and apply Min-Max normalization within each sequence to create a standardized penalty score $\hat{\ell}_{i,t} \in [0, 1]$:
\begin{equation}
\hat{\ell}_{i,t} = \frac{\ell_{i,t} - \min_{k}(\ell_{i,k})}{\max_{k}(\ell_{i,k}) - \min_{k}(\ell_{i,k}) + \epsilon}
\end{equation}
A value of $\hat{\ell}_{i,t}$ close to 1 indicates high relative certainty for that token choice. This penalty acts as a regularizer, complementing the directional guidance from Stage 1. 

\paragraph{Final Advantage Shaping Formula.}
By combining these two stages, UCAS creates a composite advantage signal that is both globally informed and locally sensitive. The final shaped advantage for each token is:
\begin{equation}
\hat{A}^{\text{UCAS}}_{i,t} =
\begin{cases}
\max(0, \hat{A}^{\text{mod}}_i - \beta \hat{\ell}_{i,t}), & \hat{A}^{\text{mod}}_i \geq 0 \\
\hat{A}^{\text{mod}}_i - \beta \hat{\ell}_{i,t}, & \hat{A}^{\text{mod}}_i < 0
\end{cases}
\label{eq:final_advantage}
\end{equation}
where $\beta > 0$ is a hyperparameter controlling the penalty strength. When $\hat{A}^{\text{mod}}_i \geq 0$, the clamp prevents correct trajectories from receiving negative token-level advantages. When $\hat{A}^{\text{mod}}_i < 0$, the penalty further suppresses high-certainty errors, assigning stronger negative updates to confidently wrong transitions. This asymmetric structure steers the model toward novel correct solutions while discouraging localized overconfidence, thereby mitigating entropy collapse and fostering more robust problem-solving abilities. This final advantage term then replaces the original advantage in the RL objective:
\begin{equation}
\begin{aligned}
\mathcal{J}_{\text{UCAS}}&(\theta) = \mathbb{E}_{\substack{(q,a) \sim \mathcal{D} \\ \{o_i\}_{i=1}^G \sim \pi_{\theta_{\text{old}}}}} \Bigg[ \frac{1}{\sum_{j=1}^G |o_j|} \\
& \quad \sum_{i=1}^G \sum_{t=1}^{|o_i|} \min \Big( r^i_t(\theta)\hat{A}^{\text{UCAS}}_{i,t}, \\
& \quad \text{clip}\big(r^i_t(\theta), 1-\epsilon_{\text{low}}, 1+\epsilon_{\text{high}}\big)\hat{A}^{\text{UCAS}}_{i,t} \Big) \Bigg] \\
\text{s.t. } & 0 < |\{ i \mid \text{is\_equiv}(o^i, a) \}| < G
\end{aligned}
\label{eq:ucas_objective}
\end{equation}
The complete implementation process of UCAS is shown in Algorithm \ref{alg:ucas_function}.

\begin{algorithm}[t]
\renewcommand{\algorithmicrequire}{\textbf{Input:}}
\renewcommand{\algorithmicensure}{\textbf{Output:}}
\caption{Uncertainty-aware Advantage Shaping (UCAS)}
\label{alg:ucas_function}
\begin{algorithmic}[1]
\Require 
\Statex A group of $G$ responses $\{o_i\}_{i=1}^G \sim \pi_\theta$;
\Statex Rule-based rewards $\{R_i\}_{i=1}^G$;
\Statex Hyperparameters $\alpha, \beta$.

\State Compute group advantages $\{\hat{A}_i\}_{i=1}^G$.
\State Compute self-confidence $\{\mathcal{C}(o_i|q)\}_{i=1}^G$.
\State Normalize confidences to get $\{\hat{\mathcal{C}}_i\}_{i=1}^G$.

\For{$i = 1$ to $G$}
    \State \textbf{\textit{Stage 1: Response Modulation}} 
    \State Calc $W(\hat{\mathcal{C}}_i)$ via Eq.~\ref{eq:modulation_weight}.
    \State $\hat{A}^{\text{mod}}_i \leftarrow W(\hat{\mathcal{C}}_i) \cdot \hat{A}_i$.
    
    \State \textbf{\textit{Stage 2: Token Certainty Penalty}}
    \State Get logits $\{\ell_{i,t}\}$ for tokens in $o_i$.
    \State Normalize logits to get $\{\hat{\ell}_{i,t}\}$.
    
    \For{$t = 1$ to $|o_i|$}
        \If{$\hat{A}^{\text{mod}}_i \geq 0$}
            \State $\hat{A}^{\text{UCAS}}_{i,t} \leftarrow \max(0, \hat{A}^{\text{mod}}_i - \beta \cdot \hat{\ell}_{i,t})$.
        \Else
            \State $\hat{A}^{\text{UCAS}}_{i,t} \leftarrow \hat{A}^{\text{mod}}_i - \beta \cdot \hat{\ell}_{i,t}$.
        \EndIf
    \EndFor
\EndFor
\Ensure Token-level advantages $\{\hat{A}^{\text{UCAS}}_{i,t}\}$.
\end{algorithmic}
\end{algorithm}

\section{Experiments}
\subsection{Experimental Setup}

\paragraph{Models and Baselines.}
We employ two variants of the Qwen2.5-Math~\citep{yang2024qwen2} series as our foundation models: Qwen2.5-Math-1.5B and Qwen2.5-Math-7B. To quantify the performance improvement introduced by our method, we compare against two widely used RLVR baselines, GRPO and DAPO. In addition, we benchmark against several representative recent methods in math reasoning and RLVR, including Simple-RL-Zoo~\citep{zeng2025simplerl}, PRIME-Zero~\citep{cui2025process}, OpenReasonerZero~\citep{hu2025open}, Oat-Zero~\citep{liu2025understanding}, GRPO with Entropy Adv.~\citep{cheng2025reasoning}, and KTAE~\citep{sun2025ktae}. Detailed descriptions of the baselines are provided in Appendix~\ref{app:baselines}.

\paragraph{Training Data and Benchmarks.}
During the training phase, we utilize the widely-used MATH dataset as our training set. To maintain consistency with prior research, we only use the more challenging subset of this dataset for training, specifically problems from levels 3 to 5. To comprehensively evaluate the reasoning capabilities of the model trained with our method, we select five widely recognized benchmarks in the mathematical reasoning domain for testing: AIME24~\citep{numina_math_datasets}, MATH-500~\citep{hendrycks2021measuring}, AMC~\citep{numina_math_datasets}, Minerva~\citep{lewkowycz2022solving}, and OlympiadBench~\citep{huang2024olympicarena}, which collectively contain 1,560 problems.

\subsection{Main Results}
The greedy pass@1 performance comparison between 1.5B and 7B models across five mathematical reasoning benchmarks is presented in Table~\ref{tab:math-main}. We can clearly find that the UCAS model achieved the highest performance across all five math reasoning benchmarks on both the 1.5B and 7B parameter scales. Compared with the DAPO baseline, UCAS improves the average accuracy from 41.2 to 47.3 (+6.1) on Qwen2.5-Math-1.5B and from 50.5 to 56.7 (+6.2) on Qwen2.5-Math-7B. Beyond DAPO, UCAS also surpasses strong baselines such as KTAE and Oat-Zero, with pronounced gains on challenging benchmarks including AIME24, AMC, and OlympiadBench. These results highlight the robustness and scalability of uncertainty-aware advantage shaping, demonstrating consistent benefits across model sizes and diverse reasoning tasks. Additional four-seed results in Appendix~\ref{app:seed_variance} show that these gains are stable and not driven by a single favorable run.

\begin{table*}[ht]
\centering
\resizebox{0.99 \textwidth}{!}{
\begin{tabular}{lcccccc}
\toprule
\textbf{Models} & \textbf{AIME24} & \textbf{MATH-500} & \textbf{AMC} & \textbf{Minerva} & \textbf{Olympiad} & \textbf{Avg} \\
\midrule
\multicolumn{7}{c}{\textit{Qwen2.5-Math-1.5B}}\\
\midrule
Base Model & 7.3 & 61.8 & 43.4 & 15.1 & 28.4 & 31.2 \\
GRPO & 15.6 & 76.0 & 51.8 & 22.1 & 36.3 & 40.4 \\
DAPO & 16.7 & 77.6 & 47.0 & 25.7 & 39.0 & 41.2 \\
Oat-Zero\citep{liu2025understanding} & 20.0 & 74.4 & 50.6 & 23.9 & 37.0 & 41.2 \\
KTAE\citep{sun2025ktae} & 20.0 & 77.6 & 50.6 & 29.0 & 40.0 & 43.4 \\
SEED-GRPO\citep{chen2025seed} & 23.3 & 75.4 & 50.6 & 26.8 & 41.3 & 43.5 \\
\rowcolor{gray!15} UCAS & \textbf{23.3} & \textbf{80.6} & \textbf{59.0} & \textbf{31.6} & \textbf{42.1} & \textbf{47.3} \\
\midrule
\multicolumn{7}{c}{\textit{Qwen2.5-Math-7B}}\\
\midrule
Base Model & 11.0 & 69.0 & 45.8 & 21.3 & 28.4 & 35.1 \\
GRPO & 30.0 & 81.0 & 57.8 & 32.7 & 43.2 & 48.9 \\
DAPO & 30.5 & 81.8 & 60.2 & 34.5 & 45.3 & 50.5 \\
PRIME-Zero \citep{cui2025process} & 23.3 & 82.2 & 57.8 & 36.0 & 39.9 & 47.8 \\
OpenReasonerZero \citep{hu2025open}  & 17.9 &  78.4 &45.8  & 27.9 & 45.0 & 43.0\\
Oat-Zero\citep{liu2025understanding} & 32.1 & 79.8 & 61.4 & 30.5 & 41.8 & 49.1 \\
Simple RL-Zero\citep{zeng2025simplerl} & 26.7 & 78.6 & 59.0 & 33.8 & 43.4 & 48.3 \\
GRPO with Entropy Adv. \citep{cheng2025reasoning}$^{\dag}$ & 33.7 & 83.1 & \textbf{69.8} & - & - & -\\
KTAE\citep{sun2025ktae} & 36.7 & 83.2 & 63.9 & 35.3 & 43.7 & 52.6 \\
SEED-GRPO\citep{chen2025seed} & 43.3 & 82.2 & 64.7 & 35.0 & 45.2 & 54.7 \\
\rowcolor{gray!15} UCAS & \textbf{43.3} & \textbf{85.6} & 68.7 & \textbf{37.6} & \textbf{48.0} & \textbf{56.7} \\
\bottomrule
\end{tabular}
}
\caption{The greedy pass@1 performance of 1.5B and 7B models across five math reasoning benchmarks. \dag :~results from \citet{cheng2025reasoning}. Our method UCAS consistently surpasses all baselines in both parameter scales.}
\label{tab:math-main}
\end{table*}

\subsection{Analysis}

\begin{table*}[h]
\centering
\resizebox{0.95 \textwidth}{!}{
\begin{tabular}{lcccccc}
\toprule
\textbf{Models} & \textbf{AIME24} & \textbf{MATH-500} & \textbf{AMC} & \textbf{Minerva} & \textbf{Olympiad} & \textbf{Avg} \\
\midrule
\multicolumn{7}{c}{\textit{Qwen2.5-Math-1.5B}}\\
\midrule
Base Model & 7.3 & 61.8 & 43.4 & 15.1 & 28.4 & 31.2 \\
w/ DAPO & 16.7 & 77.6 & 47.0 & 25.7 & 39.0 & 41.2 \\
w/ DAPO + Response-Level Confidence  & \textbf{23.3} & 79.6 & 51.8 & 27.6 & 41.0 & 44.7 \\
w/ DAPO + Token-Level Certainty & 20.0 & 80.2 & 55.4 & 29.7 & 40.1 & 45.1 \\
\rowcolor{gray!15} w/ DAPO + UCAS (Ours) & \textbf{23.3} & \textbf{80.6} & \textbf{59.0} & \textbf{31.6} & \textbf{42.1} & \textbf{47.3} \\
\midrule
\multicolumn{7}{c}{\textit{Qwen2.5-Math-7B}}\\
\midrule
Base Model & 11.0 & 69.0 & 45.8 & 21.3 & 28.4 & 35.2 \\
w/ DAPO & 30.5 & 81.8 & 60.2 & 34.5 & 45.3 & 50.5 \\
w/ DAPO + Response-Level Confidence & 40.0 & 85.0 & 63.9 & 36.7 & 47.4 & 54.6 \\
w/ DAPO + Token-Level Certainty & 36.7 & 84.6 & 65.0 & 29.7 & 47.7 & 52.7 \\
\rowcolor{gray!15} w/ DAPO + UCAS (Ours) & \textbf{43.3} & \textbf{85.6} & \textbf{68.7} & \textbf{37.6} & \textbf{48.0} & \textbf{56.7} \\
\bottomrule
\end{tabular}
}
\caption{Ablation study of uncertainty modeling. Both sentence-level and token-level uncertainty bring consistent gains over the DAPO baseline.}
\label{tab:ablation-uncertainty}
\end{table*}

\paragraph{Ablation Study.}The ablation comparison between response-level and token-level uncertainty modeling is presented in Table \ref{tab:ablation-uncertainty}. We can clearly observe that both response-level and token-level uncertainty bring consistent gains over the DAPO baseline. Compared with the model trained with DAPO, incorporating response-level confidence increases the average score on Qwen2.5-Math-1.5B from 41.2 to 44.7 (+3.5\%), while token-level uncertainty further raises it to 45.1 (+3.9\%). A similar trend holds on the 7B model, where both variants surpass the DAPO baseline.
Their integration in UCAS achieves the best performance, confirming that both signals are individually useful and jointly necessary.

\begin{table}[h]
    \centering
    \resizebox{0.49\textwidth}{!}{
    \begin{tabular}{ccc}
        \toprule
        \textbf{Response-Level $\alpha$} & \textbf{Token-Level $\beta$} & \textbf{Math-500} \\
        \midrule
        \multicolumn{3}{c}{\textit{Varying Token-Level Penalty ($\beta$), Fixed $\alpha=0.2$}} \\
        \midrule
        0.2 & 0.005 & 79.6 \\
        \rowcolor{gray!15} 0.2 & 0.01 & \textbf{80.4} \\
        0.2 & 0.05 & 78.8 \\
        0.2 & 0.1 & 78.4 \\
        \midrule
        \multicolumn{3}{c}{\textit{Varying Response Modulation ($\alpha$), Fixed $\beta=0.01$}} \\
        \midrule
        0.1 & 0.01 & 80.0 \\
        \rowcolor{gray!15} 0.2 & 0.01 & \textbf{80.4} \\
        0.4 & 0.01 & 78.8 \\
        \bottomrule
    \end{tabular}
    }
    \caption{Sensitivity analysis on Math-500 when varying the token-level penalty $\beta$ (upper block) and the response-level modulation coefficient $\alpha$ (lower block).}
    \label{tab:hyper_sensitivity}
\end{table}
\paragraph{Hyperparameter Sensitivity.}
We evaluate the robustness of UCAS by varying the response-level modulation $\alpha$ and token-level penalty $\beta$ on the Math-500 benchmark (Table~\ref{tab:hyper_sensitivity}). This stability is supported by strict normalization throughout the method: group advantages and response-level confidence are z-score normalized, while token logits are min--max normalized to $[0,1]$. A moderate penalty ($\beta = 0.01$) yields optimal performance. Lower values (0.005) fail to sufficiently counteract entropy collapse, while excessive penalization ($\beta \geq 0.05$) over-regularizes the policy. This suggests that high $\beta$ artificially flattens the distribution even for necessary, high-certainty steps, hindering coherent reasoning chains. The method exhibits stability within $\alpha \in [0.1, 0.2]$. However, aggressive modulation ($\alpha=0.4$) causes a performance drop. This implies a "signal dominance" issue: overly strong confidence scaling overshadows the fundamental correctness signal, introducing variance that distracts from the primary objective of mathematical accuracy. Notably, the same $(\alpha, \beta)$ setting also transfers effectively across the 1.5B and 7B models in Table~\ref{tab:math-main}, indicating limited sensitivity to moderate model-scale variation.

\begin{figure*}[htbp]
\begin{center}
\includegraphics[width=1.0\linewidth]{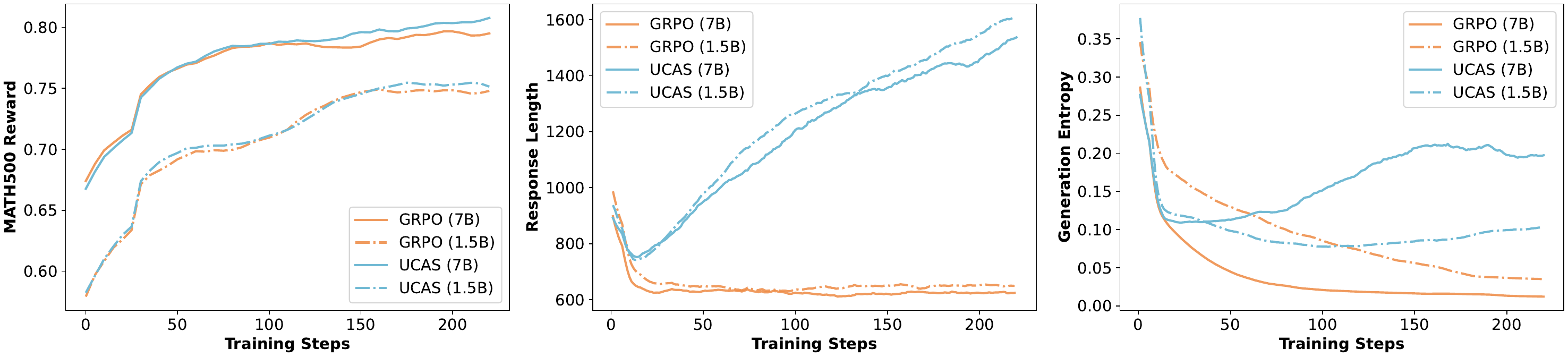}
\end{center}
\caption{Training dynamics of UCAS compared with GRPO across both 7B and 1.5B models. \textbf{Left:} Reward; \textbf{Middle:} Response Length; \textbf{Right:} Generation Entropy.}
 \label{fig:training}
\end{figure*}
\paragraph{Training Dynamics.}The training process highlights several key performance trends, as shown in Figure \ref{fig:training}. Compared to vanilla GRPO, UCAS demonstrates a consistent increase in the inference reward on the MATH500 benchmark. Regarding the average response length, the inclusion of UCAS enables the model to generate longer reasoning chains, reflecting more comprehensive problem-solving \citep{guo2025deepseek,cheng2025reasoning}, while simultaneously improving accuracy. For generation entropy, UCAS shows an early decline but later recovers and stabilizes at a higher level, effectively avoiding the entropy collapse reported in prior work \citep{cui2025entropy}. Notably, the model's reward continues to rise even as the entropy increases, which indicates a stable and effective training dynamic where exploration and optimization are well-balanced.

\begin{figure}[htbp]
\begin{center}
\includegraphics[width=1.0\linewidth]{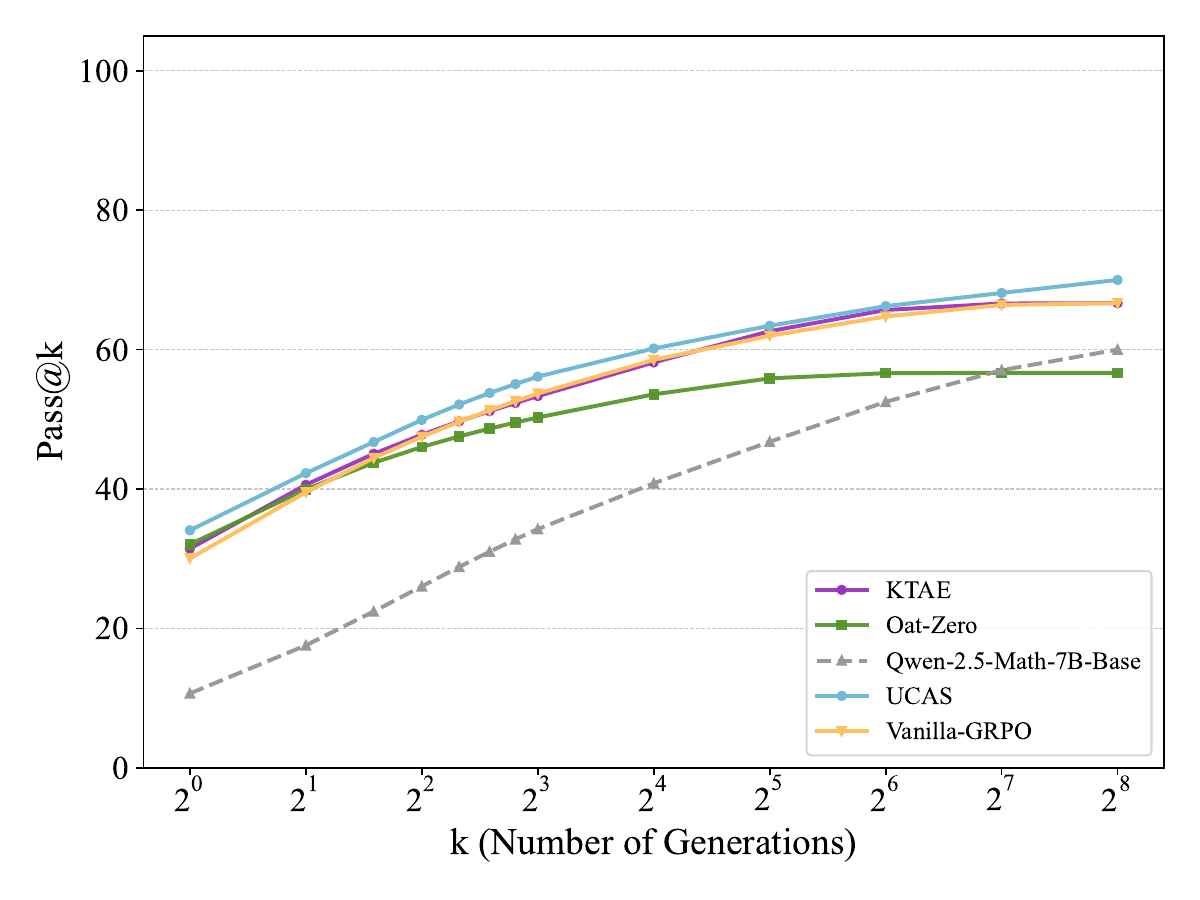}
\end{center}
\caption{Comparison of pass@k results on the AIME24 Benchmark.}
 \label{fig:passk}
\end{figure}

\paragraph{Pass@k Evaluation.}
Prior studies \citep{wang2022self,wu2024inference} have shown that with a limited number of rollouts, models often struggle to solve certain tasks. In contrast, when the rollout budget is sufficiently large, the probability of sampling effective solutions increases considerably. This observation suggests that pass@k accuracy with a large k provides a more reliable estimate of a model’s potential performance \citep{yue2025does}. Under this evaluation protocol, a problem is considered solved if any of the k sampled reasoning trajectories yield the correct answer. Figure \ref{fig:passk} reports pass@k results on the AIME24 benchmark. The results indicate that UCAS achieves more consistent improvements as k grows. In contrast, Vanilla-GRPO and its enhanced variants show slower growth, consistent with findings from \citet{yue2025does}. The stronger performance of UCAS under the pass@k metric highlights its effectiveness, which can be attributed to differences in exploration strategies. Unlike Vanilla-GRPO, which often suffers from exploration stagnation, where the model repeatedly samples low-diversity rollouts, UCAS leverages uncertainty-aware advantage shaping to sustain diverse exploration and escape local optima. Additional pass@k results on AMC and MATH-500, reported in Appendix~\ref{app:additional_passk}, show the same trend on broader benchmarks.

\section{Related Work}
\subsection{RL for LLM Reasoning}

Recent advances in reinforcement learning have transformed the training of large language models for reasoning tasks. Process reward models (PRMs) \citep{lightman2023let, nie2026attnpo, wu2026steppotentialadvantageestimation} have emerged as a key innovation, providing step-level supervision that improves both efficiency and accuracy compared to outcome-only rewards. Approaches such as PRIME \citep{cui2025process} eliminate costly human annotation by deriving implicit process feedback, while OmegaPRM \citep{luo2024improve} leverages Monte Carlo Tree Search (MCTS) to automatically identify reasoning errors. Alongside this, DeepSeek-R1 \citep{guo2025deepseek} demonstrates that sophisticated reasoning can emerge purely from RL without supervised fine-tuning, enabled by GRPO, which replaces value functions with group-based baselines. These advances redefine alignment and reasoning in LLMs, positioning reinforcement learning with verifiable or process-level rewards as a scalable and principled alternative to preference-model-based RLHF.

\subsection{Reinforcement Learning from Verifiable Rewards}
RLVR has emerged as a scalable alternative to preference-based alignment by converting open-ended outputs into checkable signals such as mathematical correctness or unit-test pass rates \citep{guo2025deepseek, yue2025does, an2025amobenchlargelanguagemodels, zhou2026look}. While early implementations demonstrated strong gains in pass@1 accuracy, subsequent studies revealed a consistent policy entropy collapse: models rapidly concentrate probability mass on a narrow set of high-reward trajectories, diminishing output diversity and limiting exploration \citep{cui2025entropy, hao2025rethinking}. Empirical analyses show that RLVR-trained models often underperform base models on pass@$k$ \citep{shao2024deepseekmath,yue2025does}, highlighting a precision--diversity trade-off \citep{wu2025invisible, dong2025rl, zhao2026reinforcedefficientreasoningsemantically}. 

Algorithmic responses to entropy collapse vary. Standard entropy or KL penalties provide partial remedies, though their effectiveness often depends heavily on the divergence form \citep{li2025choice}. More recent uncertainty-aware approaches have sought to refine the learning signal, though with differing philosophies. For instance, SEED-GRPO \citep{chen2025seed} leverages semantic entropy to downscale updates for uncertain queries, adopting a conservative risk-mitigation strategy. In stark contrast, UCAS adopts an exploratory philosophy: we explicitly amplify rewards for correct-but-uncertain trajectories to incentivize venturing into novel reasoning domains, rather than inhibiting learning from uncertainty.

Similarly, while entropy-based shaping methods \citep{cheng2025reasoning} introduce indiscriminate entropy bonuses to encourage diversity, UCAS implements a \emph{conditional}, two-stage mechanism. By combining a logit-space response-level confidence proxy with token-level raw logits, which we find to be a more sensitive signal of local overconfidence than post-softmax entropy, UCAS distinguishes between productive exploration and blind guessing. Unlike pure entropy-based frameworks, UCAS introduces correctness-contingent modulation, amplifying penalties for confident errors while guiding exploration through uncertainty, offering a more fine-grained solution to the entropy collapse problem than global regularization or token-level covariance control \citep{cui2025entropy}.

\section{Conclusion}
In this work, we introduced UnCertainty-aware Advantage Shaping (UCAS), a fine-grained advantage estimation framework that leverages internal confidence signals without requiring additional reward models. By jointly modeling uncertainty at both the response and token levels, UCAS reshapes advantages to highlight critical uncertain reasoning steps, suppress overconfident errors, and preserve non-negative token-level updates on positive trajectories. Experimental results on major mathematical reasoning benchmarks show that UCAS achieves substantial performance improvements over GRPO and its enhanced variants. Analysis of the training dynamics further reveals that, as training progresses, UCAS demonstrates steadily increasing rewards, longer reasoning chains, and an entropy trajectory that first declines and then rises, reflecting stronger exploratory capability. These findings indicate that uncertainty-aware advantage shaping offers an effective pathway toward more robust reinforcement learning for large language models.

\section*{Limitations}
Although UCAS demonstrates significant improvements in reasoning capabilities and exploration efficiency, we acknowledge several limitations that identify clear directions for future research. First, our experiments are exclusively focused on 1.5B and 7B parameter models, and the performance of UCAS on largerscale models has not yet been fully verified due to limitations of computing resources. Second, our method relies on self-confidence and raw logits as proxies for model uncertainty. While these internal signals are computationally efficient and effective, future work could explore alternative or complementary uncertainty metrics. Techniques such as Monte Carlo dropout, model ensembles, or semantic entropy could potentially capture different facets of model uncertainty and lead to even more refined and robust advantage shaping. Investigating these areas will be essential for understanding the broader generalizability of our approach.

\bibliography{custom}

@article{hendrycks2021measuring,
  title={Measuring mathematical problem solving with the math dataset},
  author={Hendrycks, Dan and Burns, Collin and Kadavath, Saurav and Arora, Akul and Basart, Steven and Tang, Eric and Song, Dawn and Steinhardt, Jacob},
  journal={arXiv preprint arXiv:2103.03874},
  year={2021}
}

@article{lewkowycz2022solving,
  title={Solving quantitative reasoning problems with language models},
  author={Lewkowycz, Aitor and Andreassen, Anders and Dohan, David and Dyer, Ethan and Michalewski, Henryk and Ramasesh, Vinay and Slone, Ambrose and Anil, Cem and Schlag, Imanol and Gutman-Solo, Theo and others},
  journal={Advances in neural information processing systems},
  volume={35},
  pages={3843--3857},
  year={2022}
}

@article{huang2024olympicarena,
  title={Olympicarena: Benchmarking multi-discipline cognitive reasoning for superintelligent ai},
  author={Huang, Zhen and Wang, Zengzhi and Xia, Shijie and Li, Xuefeng and Zou, Haoyang and Xu, Ruijie and Fan, Run-Ze and Ye, Lyumanshan and Chern, Ethan and Ye, Yixin and others},
  journal={Advances in Neural Information Processing Systems},
  volume={37},
  pages={19209--19253},
  year={2024}
}

@misc{numina_math_datasets,
  author = {Jia LI and Edward Beeching and Lewis Tunstall and Ben Lipkin and Roman Soletskyi and Shengyi Costa Huang and Kashif Rasul and Longhui Yu and Albert Jiang and Ziju Shen and Zihan Qin and Bin Dong and Li Zhou and Yann Fleureau and Guillaume Lample and Stanislas Polu},
  title = {NuminaMath},
  year = {2024},
  publisher = {Numina},
  journal = {Hugging Face repository},
  howpublished = {\url{[https://huggingface.co/AI-MO/NuminaMath-CoT](https://github.com/project-numina/aimo-progress-prize/blob/main/report/numina_dataset.pdf)}}
}

@article{yue2025does,
  title={Does reinforcement learning really incentivize reasoning capacity in llms beyond the base model?},
  author={Yue, Yang and Chen, Zhiqi and Lu, Rui and Zhao, Andrew and Wang, Zhaokai and Song, Shiji and Huang, Gao},
  journal={arXiv preprint arXiv:2504.13837},
  year={2025}
}

@article{liu2025understanding,
  title={Understanding r1-zero-like training: A critical perspective},
  author={Liu, Zichen and Chen, Changyu and Li, Wenjun and Qi, Penghui and Pang, Tianyu and Du, Chao and Lee, Wee Sun and Lin, Min},
  journal={arXiv preprint arXiv:2503.20783},
  year={2025}
}

@article{hu2025open,
  title={Open-reasoner-zero: An open source approach to scaling up reinforcement learning on the base model},
  author={Hu, Jingcheng and Zhang, Yinmin and Han, Qi and Jiang, Daxin and Zhang, Xiangyu and Shum, Heung-Yeung},
  journal={arXiv preprint arXiv:2503.24290},
  year={2025}
}

@article{sheng2024hybridflow,
  title   = {HybridFlow: A Flexible and Efficient RLHF Framework},
  author  = {Guangming Sheng and Chi Zhang and Zilingfeng Ye and Xibin Wu and Wang Zhang and Ru Zhang and Yanghua Peng and Haibin Lin and Chuan Wu},
  year    = {2024},
  journal = {arXiv preprint arXiv: 2409.19256}
}

@article{li2025choice,
  title={The Choice of Divergence: A Neglected Key to Mitigating Diversity Collapse in Reinforcement Learning with Verifiable Reward},
  author={Li, Long and Hao, Jiaran and Liu, Jason Klein and Zhou, Zhijian and Tan, Xiaoyu and Chu, Wei and Wang, Zhe and Pan, Shirui and Qu, Chao and Qi, Yuan},
  journal={arXiv preprint arXiv:2509.07430},
  year={2025}
}

@article{wu2025invisible,
  title={The invisible leash: Why rlvr may not escape its origin},
  author={Wu, Fang and Xuan, Weihao and Lu, Ximing and Harchaoui, Zaid and Choi, Yejin},
  journal={arXiv preprint arXiv:2507.14843},
  year={2025}
}

@article{dong2025rl,
  title={Rl-plus: Countering capability boundary collapse of llms in reinforcement learning with hybrid-policy optimization},
  author={Dong, Yihong and Jiang, Xue and Tao, Yongding and Liu, Huanyu and Zhang, Kechi and Mou, Lili and Cao, Rongyu and Ma, Yingwei and Chen, Jue and Li, Binhua and others},
  journal={arXiv preprint arXiv:2508.00222},
  year={2025}
}

@article{zeng2025simplerl,
  title={Simplerl-zoo: Investigating and taming zero reinforcement learning for open base models in the wild},
  author={Zeng, Weihao and Huang, Yuzhen and Liu, Qian and Liu, Wei and He, Keqing and Ma, Zejun and He, Junxian},
  journal={arXiv preprint arXiv:2503.18892},
  year={2025}
}

@article{sun2025ktae,
  title={KTAE: A Model-Free Algorithm to Key-Tokens Advantage Estimation in Mathematical Reasoning},
  author={Sun, Wei and Yang, Wen and Jian, Pu and Du, Qianlong and Cui, Fuwei and Ren, Shuo and Zhang, Jiajun},
  journal={arXiv preprint arXiv:2505.16826},
  year={2025}
}

@article{yang2024qwen2,
  title={Qwen2. 5-math technical report: Toward mathematical expert model via self-improvement},
  author={Yang, An and Zhang, Beichen and Hui, Binyuan and Gao, Bofei and Yu, Bowen and Li, Chengpeng and Liu, Dayiheng and Tu, Jianhong and Zhou, Jingren and Lin, Junyang and others},
  journal={arXiv preprint arXiv:2409.12122},
  year={2024}
}

@article{guo2025deepseek,
  title={Deepseek-r1: Incentivizing reasoning capability in llms via reinforcement learning},
  author={Guo, Daya and Yang, Dejian and Zhang, Haowei and Song, Junxiao and Zhang, Ruoyu and Xu, Runxin and Zhu, Qihao and Ma, Shirong and Wang, Peiyi and Bi, Xiao and others},
  journal={arXiv preprint arXiv:2501.12948},
  year={2025}
}

@article{luo2024improve,
  title={Improve mathematical reasoning in language models by automated process supervision},
  author={Luo, Liangchen and Liu, Yinxiao and Liu, Rosanne and Phatale, Samrat and Guo, Meiqi and Lara, Harsh and Li, Yunxuan and Shu, Lei and Zhu, Yun and Meng, Lei and others},
  journal={arXiv preprint arXiv:2406.06592},
  year={2024}
}

@article{cui2025process,
  title={Process reinforcement through implicit rewards},
  author={Cui, Ganqu and Yuan, Lifan and Wang, Zefan and Wang, Hanbin and Li, Wendi and He, Bingxiang and Fan, Yuchen and Yu, Tianyu and Xu, Qixin and Chen, Weize and others},
  journal={arXiv preprint arXiv:2502.01456},
  year={2025}
}

@inproceedings{
lightman2024lets,
title={Let's Verify Step by Step},
author={Hunter Lightman and Vineet Kosaraju and Yuri Burda and Harrison Edwards and Bowen Baker and Teddy Lee and Jan Leike and John Schulman and Ilya Sutskever and Karl Cobbe},
booktitle={The Twelfth International Conference on Learning Representations},
year={2024},
url={https://openreview.net/forum?id=v8L0pN6EOi}
}

@article{chen2024step,
  title={Step-level value preference optimization for mathematical reasoning},
  author={Chen, Guoxin and Liao, Minpeng and Li, Chengxi and Fan, Kai},
  journal={arXiv preprint arXiv:2406.10858},
  year={2024}
}

@article{sun2025efficient,
  title={An Efficient and Precise Training Data Construction Framework for Process-supervised Reward Model in Mathematical Reasoning},
  author={Sun, Wei and Du, Qianlong and Cui, Fuwei and Zhang, Jiajun},
  journal={arXiv preprint arXiv:2503.02382},
  year={2025}
}

@article{wang2023math,
  title={Math-shepherd: Verify and reinforce llms step-by-step without human annotations},
  author={Wang, Peiyi and Li, Lei and Shao, Zhihong and Xu, RX and Dai, Damai and Li, Yifei and Chen, Deli and Wu, Yu and Sui, Zhifang},
  journal={arXiv preprint arXiv:2312.08935},
  year={2023}
}

@article{zhang2024generative,
  title={Generative verifiers: Reward modeling as next-token prediction},
  author={Zhang, Lunjun and Hosseini, Arian and Bansal, Hritik and Kazemi, Mehran and Kumar, Aviral and Agarwal, Rishabh},
  journal={arXiv preprint arXiv:2408.15240},
  year={2024}
}

@article{chen2025seed,
  title={Seed-grpo: Semantic entropy enhanced grpo for uncertainty-aware policy optimization},
  author={Chen, Minghan and Chen, Guikun and Wang, Wenguan and Yang, Yi},
  journal={arXiv preprint arXiv:2505.12346},
  year={2025}
}

@article{cheng2025reasoning,
  title={Reasoning with exploration: An entropy perspective},
  author={Cheng, Daixuan and Huang, Shaohan and Zhu, Xuekai and Dai, Bo and Zhao, Wayne Xin and Zhang, Zhenliang and Wei, Furu},
  journal={arXiv preprint arXiv:2506.14758},
  year={2025}
}

@article{wang2025beyond,
  title={Beyond the 80/20 rule: High-entropy minority tokens drive effective reinforcement learning for llm reasoning},
  author={Wang, Shenzhi and Yu, Le and Gao, Chang and Zheng, Chujie and Liu, Shixuan and Lu, Rui and Dang, Kai and Chen, Xionghui and Yang, Jianxin and Zhang, Zhenru and others},
  journal={arXiv preprint arXiv:2506.01939},
  year={2025}
}

@article{yu2025dapo,
  title={Dapo: An open-source llm reinforcement learning system at scale},
  author={Yu, Qiying and Zhang, Zheng and Zhu, Ruofei and Yuan, Yufeng and Zuo, Xiaochen and Yue, Yu and Dai, Weinan and Fan, Tiantian and Liu, Gaohong and Liu, Lingjun and others},
  journal={arXiv preprint arXiv:2503.14476},
  year={2025}
}

@article{shao2024deepseekmath,
  title={Deepseekmath: Pushing the limits of mathematical reasoning in open language models},
  author={Shao, Zhihong and Wang, Peiyi and Zhu, Qihao and Xu, Runxin and Song, Junxiao and Bi, Xiao and Zhang, Haowei and Zhang, Mingchuan and Li, YK and Wu, Yang and others},
  journal={arXiv preprint arXiv:2402.03300},
  year={2024}
}

@article{wang2025reinforcement,
  title={Reinforcement learning for reasoning in large language models with one training example},
  author={Wang, Yiping and Yang, Qing and Zeng, Zhiyuan and Ren, Liliang and Liu, Liyuan and Peng, Baolin and Cheng, Hao and He, Xuehai and Wang, Kuan and Gao, Jianfeng and others},
  journal={arXiv preprint arXiv:2504.20571},
  year={2025}
}

@article{cui2025entropy,
  title={The entropy mechanism of reinforcement learning for reasoning language models},
  author={Cui, Ganqu and Zhang, Yuchen and Chen, Jiacheng and Yuan, Lifan and Wang, Zhi and Zuo, Yuxin and Li, Haozhan and Fan, Yuchen and Chen, Huayu and Chen, Weize and others},
  journal={arXiv preprint arXiv:2505.22617},
  year={2025}
}

@article{su2025crossing,
  title={Crossing the Reward Bridge: Expanding RL with Verifiable Rewards Across Diverse Domains},
  author={Su, Yi and Yu, Dian and Song, Linfeng and Li, Juntao and Mi, Haitao and Tu, Zhaopeng and Zhang, Min and Yu, Dong},
  journal={arXiv preprint arXiv:2503.23829},
  year={2025}
}

@inproceedings{lightman2023let,
  title={Let's verify step by step},
  author={Lightman, Hunter and Kosaraju, Vineet and Burda, Yuri and Edwards, Harrison and Baker, Bowen and Lee, Teddy and Leike, Jan and Schulman, John and Sutskever, Ilya and Cobbe, Karl},
  booktitle={The Twelfth International Conference on Learning Representations},
  year={2023}
}

@inproceedings{liu2025uncertainty,
  title={Uncertainty quantification and confidence calibration in large language models: A survey},
  author={Liu, Xiaoou and Chen, Tiejin and Da, Longchao and Chen, Chacha and Lin, Zhen and Wei, Hua},
  booktitle={Proceedings of the 31st ACM SIGKDD Conference on Knowledge Discovery and Data Mining V. 2},
  pages={6107--6117},
  year={2025}
}

@article{ma2025estimating,
  title={Estimating llm uncertainty with logits},
  author={Ma, Huan and Chen, Jingdong and Wang, Guangyu and Zhang, Changqing},
  journal={arXiv e-prints},
  pages={arXiv--2502},
  year={2025}
}

@article{wang2022self,
  title={Self-consistency improves chain of thought reasoning in language models},
  author={Wang, Xuezhi and Wei, Jason and Schuurmans, Dale and Le, Quoc and Chi, Ed and Narang, Sharan and Chowdhery, Aakanksha and Zhou, Denny},
  journal={arXiv preprint arXiv:2203.11171},
  year={2022}
}

@article{wu2024inference,
  title={Inference scaling laws: An empirical analysis of compute-optimal inference for problem-solving with language models},
  author={Wu, Yangzhen and Sun, Zhiqing and Li, Shanda and Welleck, Sean and Yang, Yiming},
  journal={arXiv preprint arXiv:2408.00724},
  year={2024}
}

@article{jain2024livecodebench,
  title={Livecodebench: Holistic and contamination free evaluation of large language models for code},
  author={Jain, Naman and Han, King and Gu, Alex and Li, Wen-Ding and Yan, Fanjia and Zhang, Tianjun and Wang, Sida and Solar-Lezama, Armando and Sen, Koushik and Stoica, Ion},
  journal={arXiv preprint arXiv:2403.07974},
  year={2024}
}

@article{guo2024deepseek,
  title={DeepSeek-Coder: When the Large Language Model Meets Programming--The Rise of Code Intelligence},
  author={Guo, Daya and Zhu, Qihao and Yang, Dejian and Xie, Zhenda and Dong, Kai and Zhang, Wentao and Chen, Guanting and Bi, Xiao and Wu, Yu and Li, YK and others},
  journal={arXiv preprint arXiv:2401.14196},
  year={2024}
}

@article{hendrycks2020measuring,
  title={Measuring massive multitask language understanding},
  author={Hendrycks, Dan and Burns, Collin and Basart, Steven and Zou, Andy and Mazeika, Mantas and Song, Dawn and Steinhardt, Jacob},
  journal={arXiv preprint arXiv:2009.03300},
  year={2020}
}

@inproceedings{
li2026mol,
title={MoL: Adaptive Mixture-of-Length Reasoning for Efficient Question Answering with Context},
author={Guocong Li and Jinjian Zhang and Ping Wang and Dongnan Liu and Tian Liang and Qiuyi Qi and Hao Huang and Siyan Guo and Mutian Bao and Wei Zhou and Linjian Mo and Hongxia Xu and Jian Wu},
booktitle={The Fourteenth International Conference on Learning Representations},
year={2026},
url={https://openreview.net/forum?id=oWWAeLEdE3}
}

@article{hao2025rethinking,
  title={Rethinking entropy interventions in rlvr: An entropy change perspective},
  author={Hao, Zhezheng and Wang, Hong and Liu, Haoyang and Luo, Jian and Yu, Jiarui and Dong, Hande and Lin, Qiang and Wang, Can and Chen, Jiawei},
  journal={arXiv preprint arXiv:2510.10150},
  year={2025}
}

@article{nie2026attnpo,
  title={ATTNPO: Attention-Guided Process Supervision for Efficient Reasoning},
  author={Nie, Shuaiyi and Ding, Siyu and Zhang, Wenyuan and Yu, Linhao and Yang, Tianmeng and Chen, Yao and Liu, Tingwen and Yin, Weichong and Sun, Yu and Wu, Hua},
  journal={arXiv preprint arXiv:2602.09953},
  year={2026}
}

@misc{wu2026steppotentialadvantageestimation,
      title={Step Potential Advantage Estimation: Harnessing Intermediate Confidence and Correctness for Efficient Mathematical Reasoning}, 
      author={Fei Wu and Zhenrong Zhang and Qikai Chang and Jianshu Zhang and Quan Liu and Jun Du},
      year={2026},
      eprint={2601.03823},
      archivePrefix={arXiv},
      primaryClass={cs.CL},
      url={https://arxiv.org/abs/2601.03823}, 
}

@article{zhou2026look,
  title={Look Inward to Explore Outward: Learning Temperature Policy from LLM Internal States via Hierarchical RL},
  author={Zhou, Yixiao and Li, Yang and Cheng, Dongzhou and Fan, Hehe and Cheng, Yu},
  journal={arXiv preprint arXiv:2602.13035},
  year={2026}
}

@misc{zhao2026reinforcedefficientreasoningsemantically,
      title={Reinforced Efficient Reasoning via Semantically Diverse Exploration}, 
      author={Ziqi Zhao and Zhaochun Ren and Jiahong Zou and Liu Yang and Zhiwei Xu and Xuri Ge and Zhumin Chen and Xinyu Ma and Daiting Shi and Shuaiqiang Wang and Dawei Yin and Xin Xin},
      year={2026},
      eprint={2601.05053},
      archivePrefix={arXiv},
      primaryClass={cs.AI},
      url={https://arxiv.org/abs/2601.05053}, 
}

@misc{an2025amobenchlargelanguagemodels,
      title={AMO-Bench: Large Language Models Still Struggle in High School Math Competitions}, 
      author={Shengnan An and Xunliang Cai and Xuezhi Cao and Xiaoyu Li and Yehao Lin and Junlin Liu and Xinxuan Lv and Dan Ma and Xuanlin Wang and Ziwen Wang and Shuang Zhou},
      year={2025},
      eprint={2510.26768},
      archivePrefix={arXiv},
      primaryClass={cs.CL},
      url={https://arxiv.org/abs/2510.26768}, 
}

@inproceedings{lin2026cec,
  title={Cec-zero: Zero-supervision character error correction with self-generated rewards},
  author={Lin, Zhiming and Zhao, Kai and Zhang, Sophie and Yu, Peilai and Xiao, Canran},
  booktitle={Proceedings of the AAAI Conference on Artificial Intelligence},
  volume={40},
  number={28},
  pages={23612--23620},
  year={2026}
}

@misc{jiang2026foeforesterrorsmakes,
      title={FoE: Forest of Errors Makes the First Solution the Best in Large Reasoning Models}, 
      author={Kehan Jiang and Haonan Dong and Zhaolu Kang and Zhengzhou Zhu and Guojie Song},
      year={2026},
      eprint={2604.02967},
      archivePrefix={arXiv},
      primaryClass={cs.AI},
      url={https://arxiv.org/abs/2604.02967}, 
}

\clearpage
\onecolumn
\appendix
\section*{Appendix}

\startcontents[sections]
\printcontents[sections]{l}{1}{\setcounter{tocdepth}{2}}

\clearpage
\twocolumn
\section{Implementation Detials}
\subsection{Baselines}
\label{app:baselines}
\begin{itemize}
    \item[\textbullet] \textbf{Simple-RL-Zoo \citep{zeng2025simplerl}}: Based on Qwen2.5-Math-7B, trained on the math-level3-5 dataset using the standard GRPO algorithm with rule-based rewards. 
    \item[\textbullet] \textbf{PRIME-Zero \citep{cui2025process}}: An online PRM update approach that leverages implicit process rewards from rollouts and outcome labels without requiring explicit annotations. 
    \item[\textbullet] \textbf{OpenReasonerZero \citep{hu2025open}}: A zero-RL baseline on Qwen2.5-7B employing the standard PPO algorithm for policy optimization. 
    \item[\textbullet] \textbf{Oat-Zero \citep{liu2025understanding}}: Built on Qwen2.5-Math-7B, trained with rule-based rewards using a modified Dr.GRPO algorithm that removes variance terms and applies token-level normalization in the policy loss. 
    \item[\textbullet] \textbf{GRPO with Entropy Adv. \citep{cheng2025reasoning}}: Extends RLVR training by incorporating a clipped, gradient-detached entropy term into the advantage function to encourage exploration. 
    \item[\textbullet] \textbf{KTAE \citep{sun2025ktae}}: A token-level advantage estimation method trained with DAPO, quantifying key-token contributions via statistical association tests and combining them with rollout-level advantages. 
\end{itemize}
These baselines cover applications of fundamental RL algorithms, process-reward-based methods, and algorithms improved for specific tasks like mathematical reasoning, aiming to evaluate the effectiveness and novelty of our method from multiple perspectives.

\subsection{RL Training Configuration}
We adopt the VERL framework \citep{sheng2024hybridflow} and train our model using the optimization objective defined in Eq.~\ref{eq:ucas_objective}. For both GRPO and DAPO, we use the hyperparameters in Table \ref{train_hyperparameters}, without using entropy or KL losses. All experiments are conducted on 2 compute nodes, each equipped with 8*80GB GPUs.


\begin{table}[!htb]
\centering
\begin{tabular}{@{}ll@{}}
\toprule
\textbf{Hyperparameter} & \textbf{Value} \\
\midrule
Optimizer & AdamW \\
Actor learning rate & $1\text{e}^{-6}$ \\
Max prompt length & 1024 \\
Max response length & 3072 \\
Training batch size & 512 \\
Samples per prompt & 16 \\
Mini-batch size & 32  \\
Rollout temperature & 1.0 \\
Clip range $\epsilon_{\text{low}}$, $\epsilon_{\text{high}}$ & $0.2$, $0.28$ \\
UCAS hyperparameter $\alpha$, $\beta$ & $0.25$, $0.01$\\
\bottomrule
\end{tabular}
\vspace{-5pt}
\caption{RL Hyperparameters}\label{train_hyperparameters}
\end{table}

\section{Further Analysis}
\subsection{Exploratory Reasoning Dynamics}
To further analyze the effect of UCAS training, we compute the response-level confidence scores of model outputs according to Eq. \ref{eq:self_confidence}, measured before and after training on Qwen2.5-Math-1.5B across MATH and Olympiad datasets. We focus on the MATH and Olympiad datasets because they contain more samples and a larger number of responses whose correctness changes after training, which makes them well suited for detailed analysis. For comparability, the confidence values are normalized by subtracting the mean and dividing by the standard deviation. 

\begin{figure*}[ht] \centering \includegraphics[width=0.95\linewidth]{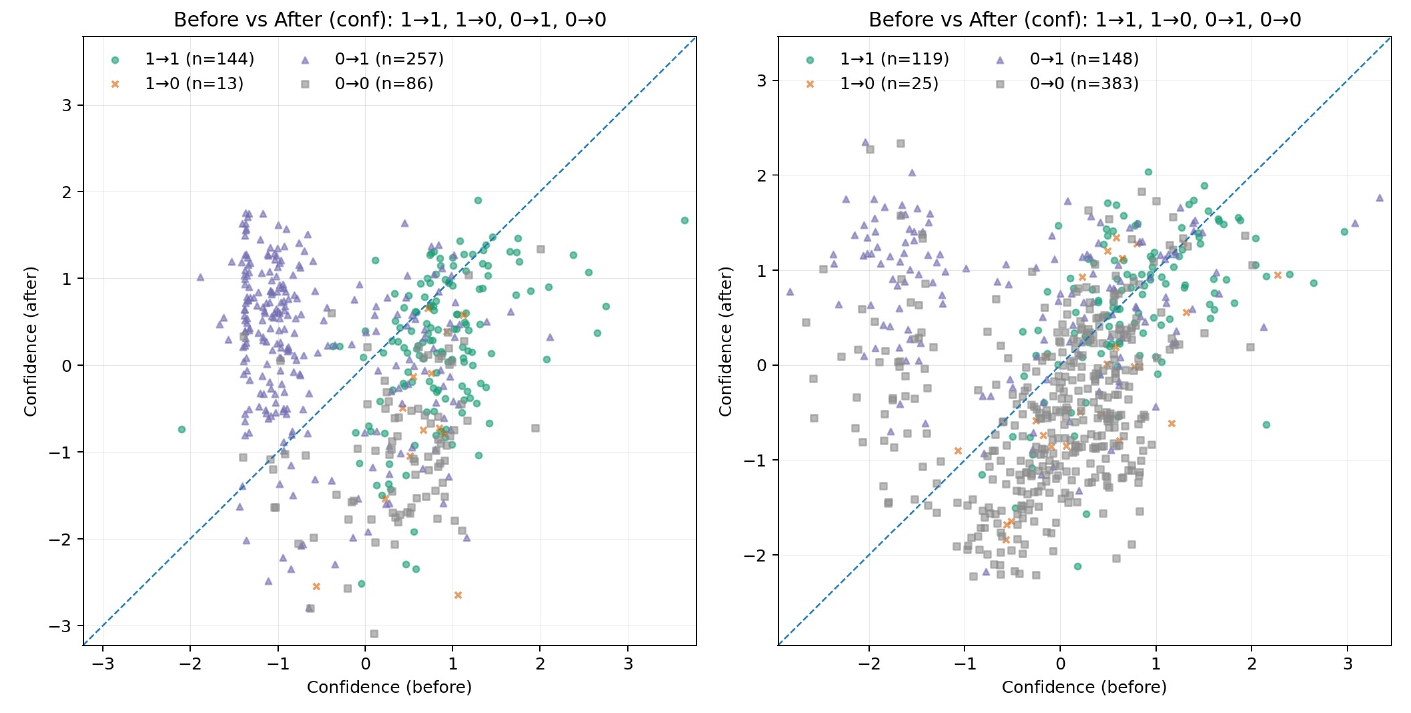} \caption{Confidence dynamics before and after UCAS training on the MATH and Olympiad datasets.} \label{fig:olympiad_bench} \end{figure*}

Based on the correctness of the responses before and after training, the samples are categorized into three groups: (i) consistently correct (1→1), (ii) correct before but incorrect after (1→0), (iii) incorrect before but correct after (0→1), and (iiii) incorrect before and incorrect after (0→0). Figure~\ref{fig:olympiad_bench} illustrates the distribution of these categories, where each point represents model's response to a given problem.
 
From Figure~\ref{fig:olympiad_bench}, we observe that for many problems correctly solved only after UCAS training (0→1), the model’s confidence notably increases. In contrast, for problems that remain unsolved before and after training (0→0), the model tends to reduce its confidence, suggesting a more calibrated estimation of its own uncertainty.

\subsection{Cross-domain Generalization}
To assess the generality of UCAS beyond mathematical reasoning, we conduct additional experiments on three diverse benchmarks: \textbf{LeetCode} \citep{guo2024deepseek} (code generation), \textbf{LiveCodeBench} \citep{jain2024livecodebench} (competitive programming), and \textbf{MMLU} \citep{hendrycks2020measuring} (general task reasoning). As shown in Table \ref{tab:general_eval}, despite being trained solely on mathematical reasoning data, UCAS consistently outperforms the strong DAPO baseline across all non-math tasks.

\begin{table}[t]
\centering
\setlength{\tabcolsep}{3.5pt}

\resizebox{\linewidth}{!}{
\begin{tabular}{lcccc}
\toprule
\textbf{Method} & \textbf{LeetCode} & \textbf{LiveCode} & \textbf{MMLU} & \textbf{Avg} \\
 & \small{(Pass@1)} & \small{(Pass@1)} & \small{(Acc)} & \\
\midrule
Base Model & 11.7 & 5.7 & 65.7 & 27.7 \\
DAPO & 18.3 & 9.2 & 67.3 & 31.6 \\
\rowcolor{gray!15} UCAS (Ours) & \textbf{23.6} \small{(+5.3)} & \textbf{14.8} \small{(+5.6)} & \textbf{70.8} \small{(+3.5)} & \textbf{36.4} \small{(+4.8)} \\
\bottomrule
\end{tabular}
}
\caption{Generalizing UCAS from math-only training to evaluations on non-math tasks.}
\label{tab:general_eval}
\end{table}

This strong transferability suggests that our uncertainty-aware exploration mechanism is a broadly applicable principle. By unlocking exploration for high-uncertainty paths, UCAS improves performance not just in calculation but also in the multi-step logical planning required for programming and general reasoning, demonstrating gains beyond the mathematical domain.

\subsection{Additional Pass@k Results}
\label{app:additional_passk}
To further substantiate the exploration and diversity claims beyond AIME24, we evaluate pass@k on AMC and MATH-500 using Qwen2.5-Math-7B. As shown in Tables~\ref{tab:passk_amc_appendix} and \ref{tab:passk_math500_appendix}, UCAS exhibits consistently stronger scaling than the DAPO baseline as the sampling budget increases. The improvement is particularly pronounced on AMC, where UCAS maintains a clear advantage across all $k$ values from 4 to 64. On MATH-500, UCAS also yields higher pass@k than DAPO, indicating that the broader search behavior induced by uncertainty-aware shaping transfers beyond a single benchmark.

\begin{table}[t]
\centering
\resizebox{\linewidth}{!}{
\begin{tabular}{lccccc}
\toprule
\textbf{Method} & \textbf{Pass@4} & \textbf{Pass@8} & \textbf{Pass@16} & \textbf{Pass@32} & \textbf{Pass@64} \\
\midrule
Base Model & 65.06 & 74.70 & 80.72 & 84.34 & 87.95 \\
DAPO & 72.56 & 82.78 & 86.64 & 88.10 & 89.50 \\
\rowcolor{gray!15} UCAS & \textbf{77.80} & \textbf{87.24} & \textbf{90.12} & \textbf{92.16} & \textbf{93.58} \\
\bottomrule
\end{tabular}
}
\caption{Additional pass@k results on AMC with Qwen2.5-Math-7B. UCAS consistently outperforms the DAPO baseline as the generation budget increases.}
\label{tab:passk_amc_appendix}
\end{table}

\begin{table}[t]
\centering
\resizebox{0.82\linewidth}{!}{
\begin{tabular}{lccc}
\toprule
\textbf{Method} & \textbf{Pass@4} & \textbf{Pass@8} & \textbf{Pass@16} \\
\midrule
Base Model & 74.40 & 82.60 & 88.80 \\
DAPO & 89.20 & 91.20 & 92.80 \\
\rowcolor{gray!15} UCAS & \textbf{91.40} & \textbf{92.10} & \textbf{93.40} \\
\bottomrule
\end{tabular}
}
\caption{Additional pass@k results on MATH-500 with Qwen2.5-Math-7B. UCAS preserves an advantage over DAPO at each sampling budget.}
\label{tab:passk_math500_appendix}
\end{table}

\subsection{Variance Across Random Seeds}
\label{app:seed_variance}
To assess the statistical reliability of the gains, we conduct four independent training and evaluation runs for both DAPO and UCAS using different random seeds on Qwen2.5-Math-7B. Table~\ref{tab:seed_variance_appendix} reports the mean and standard deviation across runs. UCAS achieves higher mean performance than DAPO on every benchmark while maintaining comparable, and in several cases smaller, standard deviations. These results indicate that the improvements reported in the main paper are not driven by a single favorable run and that the method remains stable across random initializations.

\begin{table*}[t]
\centering
\resizebox{0.93\textwidth}{!}{
\begin{tabular}{lcccccc}
\toprule
\textbf{Method} & \textbf{AIME24} & \textbf{MATH-500} & \textbf{AMC} & \textbf{Minerva} & \textbf{Olympiad} & \textbf{Avg} \\
\midrule
Qwen2.5-Math-7B (Base) & 11.0 & 69.0 & 45.8 & 21.3 & 28.4 & 35.1 \\
DAPO & $28.1_{\pm 3.9}$ & $79.4_{\pm 2.2}$ & $57.6_{\pm 3.8}$ & $32.4_{\pm 2.6}$ & $43.4_{\pm 2.2}$ & 48.18 \\
\rowcolor{gray!15} UCAS & $\mathbf{40.6}_{\pm 3.5}$ & $\mathbf{84.6}_{\pm 1.6}$ & $\mathbf{65.3}_{\pm 3.3}$ & $\mathbf{36.7}_{\pm 1.9}$ & $\mathbf{46.5}_{\pm 2.3}$ & \textbf{54.74} \\
\bottomrule
\end{tabular}
}
\caption{Performance variance across four independent trials on Qwen2.5-Math-7B. UCAS consistently improves the mean score over DAPO with comparable variance.}
\label{tab:seed_variance_appendix}
\end{table*}

\subsection{Case Study} 
As shown in Figure~\ref{fig:case_study}, the baseline model exhibits a critical failure in logical modeling. While it correctly identifies the symmetry of the hyperbola and the basic distance formula, it treats the vertices $B$ and $D$ as independent points on the hyperbola symmetric about the origin. Crucially, it fails to incorporate the \textit{rhombus constraint}, which necessitates that the diagonals $AC$ and $BD$ must be perpendicular. This omission reduces the problem to a trivial minimization of the $x$-coordinate ($x^2=20$), resulting in an incorrect lower bound of 80.

In contrast, the model trained with UCAS (Response Parts 1 and 2) demonstrates a significantly higher degree of structural awareness. Rather than performing a simple substitution, the model engages in self-reflection to identify the implicit geometric constraints of the problem. It explicitly parameterizes the diagonals as perpendicular lines and performs a domain verification to ensure they actually intersect the hyperbola. This rigorous reasoning process allows the model to discover the hidden domain of the slope parameter and use monotonicity analysis to reach the correct ground-truth answer.

The baseline's failure highlights the tendency of standard algorithms to converge on superficial "safe" trajectories that ignore implicit constraints. UCAS, by rewarding exploration of high-uncertainty paths that maintain semantic consistency, enables the model to derive complex geometric dependencies and reach the correct ground-truth value of 480 via rigorous logical deduction and code verification.

\section{Theoretical Explanation}
\label{app:theory}

In this section, we provide a theoretical analysis of the Uncertainty-aware Advantage Shaping (UCAS) method. We demonstrate that its heuristic components---exponential advantage modulation and min-max logit penalties---arise naturally from principles of risk-sensitive importance weighting and adaptive gradient regularization. This analysis establishes UCAS not as an ad-hoc collection of tricks, but as a coherent algorithmic framework for countering entropy collapse in sparse-reward RLVR.

\subsection{Problem Formulation: Entropy Collapse}
In RLVR, reward signals are typically sparse and binary. A known pathology in this setting is \textit{Entropy Collapse} \citep{cui2025entropy}, where the policy $\pi_\theta$ prematurely converges to a deterministic distribution, severely limiting exploration.
Mathematically, as the policy becomes deterministic, its entropy $H(\pi_\theta(\cdot|s)) \to 0$. Consequently, the Kullback-Leibler divergence from the policy to the uniform distribution $U(\mathcal{V})$ approaches its maximum:
\begin{equation}
    D_{\text{KL}}(\pi_\theta \| U) = \log|\mathcal{V}| - H(\pi_\theta) \to \log|\mathcal{V}|.
\end{equation}
This divergence quantifies the policy's deviation from a maximally exploratory prior. Standard RLVR algorithms lack an explicit mechanism to penalize this deviation, often converging to local optima that exploit a narrow set of high-confidence but potentially suboptimal reasoning paths.

\subsection{Stage 1: Risk-Sensitive Advantage Modulation}
\label{sec:theory_stage1}
The first stage applies a confidence-dependent modulation $W(\hat{\mathcal{C}}_i)$ to the advantage estimates. We first clarify that the response-level confidence used in UCAS is implemented entirely in logit space, and then interpret the resulting modulation as a form of risk-sensitive importance weighting.

\paragraph{Logit-Space Confidence and Its KL Interpretation.}
For token $t$ in trajectory $o_i$, let $p_{i,t}(v)=\mathrm{softmax}(\ell_{i,t})_v$ and $U(v)=1/|\mathcal{V}|$. The token-level confidence used by UCAS is
\begin{equation}
C_{i,t} = \operatorname{LSE}(\ell_{i,t}) - \frac{1}{|\mathcal{V}|}\sum_{v \in \mathcal{V}} \ell_{i,t,v}.
\end{equation}
Using $\log p_{i,t}(v)=\ell_{i,t,v}-\operatorname{LSE}(\ell_{i,t})$, we obtain
\begin{equation}
\begin{aligned}
\mathrm{KL}(U \| p_{i,t})
&= \frac{1}{|\mathcal{V}|} \sum_{v \in \mathcal{V}} \left(\log \frac{1}{|\mathcal{V}|} - \log p_{i,t}(v)\right) \\
&= \operatorname{LSE}(\ell_{i,t}) - \frac{1}{|\mathcal{V}|}\sum_{v \in \mathcal{V}} \ell_{i,t,v} \\
&\quad - \log |\mathcal{V}|.
\end{aligned}
\end{equation}
Therefore,
\begin{equation}
C_{i,t} = \mathrm{KL}(U \| p_{i,t}) + \log |\mathcal{V}|.
\end{equation}
This equivalence shows that UCAS preserves the intended uncertainty signal while avoiding direct probability-space evaluation near zero probabilities. Compared with entropy-based quantities driven by expectation under $p_{i,t}$, $\mathrm{KL}(U \| p_{i,t})$ is more sensitive to support contraction and therefore better aligned with our goal of detecting premature distributional collapse.

Let $y_i = \text{sign}(\hat{A}_i) \in \{+1, -1\}$ indicate the favorability of a trajectory. To encourage robust exploration, we seek a weighting distribution $w$ over trajectories within a mini-batch that emphasizes informative cases—specifically, those where the outcome $y_i$ and the confidence $\hat{\mathcal{C}}_i$ are \textit{anti-correlated} (e.g., correct but uncertain).

We derive this by solving for the maximum entropy distribution $w$ that shifts the expected signed confidence to a target value $\mu^*$, lower than its empirical mean under a uniform prior:
\begin{align}
    \max_{w \in \Delta^{G-1}} \quad & H(w) = - \sum_{i=1}^{G} w_i \log w_i \\
    \text{s.t.} \quad & \sum_{i=1}^{G} w_i = 1, \quad \sum_{i=1}^{G} w_i (y_i \hat{\mathcal{C}}_i) = \mu^*.
\end{align}
Solving this via Lagrange multipliers yields the Boltzmann distribution:
\begin{equation}
    w_i^* \propto \exp\big( -\alpha \cdot y_i \cdot \hat{\mathcal{C}}_i \big),
\end{equation}
where $\alpha > 0$ controls the intensity of the shift.

\paragraph{Connection to UCAS.} The modulated advantage in UCAS, $\hat{A}^{\text{mod}}_i = \hat{A}_i \cdot \exp(-\alpha y_i \hat{\mathcal{C}}_i)$, is directly proportional to this optimal importance weight $w_i^*$. Thus, Stage 1 amplifies updates from trajectories valuable for exploration (correct/uncertain) and strongly penalizes those indicative of overfitting (incorrect/confident).

\subsection{Stage 2: Adaptive Gradient Stabilization}
\label{sec:theory_stage2}
Stage 2 addresses entropy collapse at the token level via an asymmetric penalty term. We interpret this as a mechanism for adaptive gradient stabilization that suppresses confident errors while safeguarding positive trajectories.

\paragraph{Rationale for Min-Max Normalization.}
The penalty uses the Min-Max normalized logit $\hat{\ell}_{i,t} \in [0, 1]$. To ensure numerical stability and fit within column constraints, we denote the sequence-level extrema as $\ell^{\min}_i = \min_{k} \ell_{i,k}$ and $\ell^{\max}_i = \max_{k} \ell_{i,k}$. The normalized logit is defined as:
\begin{equation}
\hat{\ell}_{i,t} = \frac{\ell_{i,t} - \ell^{\min}_i}{\ell^{\max}_i - \ell^{\min}_i + \epsilon}.
\end{equation}
This design serves two critical functions:
1.  \textbf{Scale Invariance:} It removes dependence on the absolute logit scale, which varies across layers and training stages.
2.  \textbf{Relative Peak Detection:} It measures how peaked the distribution is for the chosen token \textit{relative to other steps in the same sequence}. A value of $\hat{\ell}_{i,t} \approx 1$ indicates the model is maximally confident at step $t$ compared to its temporal neighbors, signaling a local collapse point.

\paragraph{Effect on Optimization Dynamics.}
The composite advantage is
\begin{equation}
\hat{A}^{\text{UCAS}}_{i,t} =
\begin{cases}
\max(0, \hat{A}^{\text{mod}}_i - \beta \hat{\ell}_{i,t}), & \hat{A}^{\text{mod}}_i \ge 0,\\
\hat{A}^{\text{mod}}_i - \beta \hat{\ell}_{i,t}, & \hat{A}^{\text{mod}}_i < 0.
\end{cases}
\end{equation}
For unfavorable trajectories with $\hat{A}^{\text{mod}}_i < 0$, the gradient update becomes
\begin{equation}
\nabla \mathcal{J} \propto \mathbb{E} \left[ \big( \hat{A}^{\text{mod}}_i - \beta \hat{\ell}_{i,t} \big) \nabla \log \pi(o_{i,t} | s) \right].
\end{equation}
This acts as a soft, adaptive gradient damper. When $\hat{\ell}_{i,t}$ is large, the effective driving signal becomes more negative, applying stronger suppression to confidently wrong transitions and preventing aggressive probability concentration on error-prone tokens. For favorable trajectories with $\hat{A}^{\text{mod}}_i \ge 0$, the clamp ensures that the token-level penalty can attenuate positive updates but cannot reverse them. This context-aware regularization applies strong pressure only where needed while preserving non-negative learning signals for globally beneficial reasoning paths.

\subsection{Synthesized Interpretation}
Combining both stages, UCAS optimizes a policy subject to two complementary, uncertainty-aware regularizers:
\begin{equation}
\label{eq:synthesized_objective}
\begin{aligned}
\pi^* \approx \arg\max_{\pi} \Big( & \mathbb{E}_{\tau}[R(\tau)] + \lambda_1 \mathcal{R}_{\text{macro}}(\pi) \\
& + \lambda_2 \mathcal{R}_{\text{micro}}(\pi) \Big).
\end{aligned}
\end{equation}
\textbf{Macro Regularizer $\mathcal{R}_{\text{macro}}$ (Stage 1)} re-weights trajectory gradients to favor correctness with lower confidence.
\textbf{Micro Regularizer $\mathcal{R}_{\text{micro}}$ (Stage 2)} acts as an asymmetric constraint on token-level logits, suppressing confident errors while preserving non-negative updates on favorable trajectories.
This dual-level framework aligns the global objective (robust correctness) with stable local optimization dynamics.

\section{Prompt}
As shown in Figure \ref{prompt_template}, we use the same prompt template (Qwen-Math template) for both RL training and evaluation.

\begin{figure*}[t] 
    \centering
    \begin{tcolorbox}[
        colback=white,      
        colframe=black,     
        boxrule=1pt,        
        arc=2mm,            
        width=\textwidth,   
        title=\textbf{\large Prompt templates of RL training and Evaluation}, 
        coltitle=black,     
        colbacktitle=white, 
        titlerule=0mm,      
        left=2mm, right=2mm, top=2mm, bottom=2mm, 
        fonttitle=\sffamily 
    ]
        \ttfamily 
        <|im\_start|>system\textbackslash nPlease reason step by step, and put your final answer\\[1em]
        within \textbackslash\textbackslash boxed\{\}.<|im\_end|>\textbackslash n<|im\_start|>user\textbackslash n\textcolor{red}{\{question\}}<|im\_end|>\textbackslash n<|im\\[1em]
        start|>assistant\textbackslash n
    \end{tcolorbox}
    \caption{Prompt template in RL training and evaluation.}
    \label{prompt_template}
\end{figure*}

\begin{figure*}[t]
    \centering
    
    \begin{tcolorbox}[
        title={Question and Ground Truth},
        colbacktitle=headergray, 
        coltitle=white, 
        colframe=headergray
    ]
        \textbf{Question:} Let $A$, $B$, $C$, and $D$ be points on the hyperbola $\frac{x^2}{20}- \frac{y^2}{24} = 1$ such that $ABCD$ is a rhombus whose diagonals intersect at the origin. Find the greatest real number that is less than $BD^2$ for all such rhombi.
        \par\vspace{3pt}
        \textbf{Answer:} 480
    \end{tcolorbox}
    
    \vspace{4pt}

    \begin{tcolorbox}[
        title={Response from RL Baseline \hfill \textnormal},
        colbacktitle=baselineblue!80!black, 
        coltitle=white, 
        colframe=baselineblue!80!black,
        colback=baselineblue!10!white
    ]
        \scriptsize 
        To solve this problem, we need to understand the properties of the hyperbola and the rhombus. The hyperbola is given by the equation $\frac{x^2}{20} - \frac{y^2}{24} = 1$. The diagonals of the rhombus intersect at the origin, which means the rhombus is symmetric with respect to the origin. This implies that if $(x_1, y_1)$ is a point on the hyperbola, then $(-x_1, -y_1)$, $(x_1, -y_1)$, and $(-x_1, y_1)$ are also points on the hyperbola. The diagonals of the rhombus are the line segments connecting these points.

        The length of the diagonal $BD$ can be calculated using the distance formula. Since the diagonals intersect at the origin, the length of the diagonal $BD$ is the distance between $(x_1, y_1)$ and $(-x_1, -y_1)$. The distance formula gives us:
        \[ BD = \sqrt{(x_1 - (-x_1))^2 + (y_1 - (-y_1))^2} = \sqrt{(2x_1)^2 + (2y_1)^2} = 2\sqrt{x_1^2 + y_1^2}. \]
        Thus, $BD^2 = 4(x_1^2 + y_1^2)$. Since $(x_1, y_1)$ lies on the hyperbola, we have $\frac{x_1^2}{20} - \frac{y_1^2}{24} = 1$. We can express $y_1^2$ in terms of $x_1^2$:
        \[ y_1^2 = 24\left(\frac{x_1^2}{20} - 1\right) = \frac{24x_1^2}{20} - 24 = \frac{6x_1^2}{5} - 24. \]
        Substituting this into the expression for $BD^2$, we get:
        \[ BD^2 = 4\left(x_1^2 + \frac{6x_1^2}{5} - 24\right) = 4\left(\frac{5x_1^2 + 6x_1^2}{5} - 24\right) = 4\left(\frac{11x_1^2}{5} - 24\right) = \frac{44x_1^2}{5} - 96. \]
        To find the greatest real number that is less than $BD^2$ for all such rhombi, we need to find the minimum value of $BD^2$. The minimum value of $BD^2$ occurs when $x_1^2$ is minimized. Since $x_1^2$ must be greater than or equal to 20 (because $\frac{x_1^2}{20} \geq 1$), the minimum value of $x_1^2$ is 20. Substituting $x_1^2 = 20$ into the expression for $BD^2$, we get:
        \[ BD^2 = \frac{44 \cdot 20}{5} - 96 = \frac{880}{5} - 96 = 176 - 96 = 80. \]
        Therefore, the greatest real number that is less than $BD^2$ for all such rhombi is 80. Let's confirm this with Python code.
\begin{lstlisting}
# Define the function for BD^2 in terms of x1^2
def bd_squared(x1_squared):
    return (44 * x1_squared) / 5 - 96

# The minimum value of x1^2 is 20
x1_squared_min = 20
bd_squared_min = bd_squared(x1_squared_min)
print(bd_squared_min)
\end{lstlisting}
\begin{lstlisting}[backgroundcolor=\color{white}, frame=none]
80.0
\end{lstlisting}
        The greatest real number that is less than $BD^2$ for all such rhombi is \(\boxed{80}\).
    \end{tcolorbox}
    
    \caption{\textbf{Reasoning Case Study.} The baseline model (above) fails to account for the perpendicular constraint of the rhombus diagonals, reducing the problem towards a wrong path.}
    \label{fig:case_study} 
\end{figure*}

\begin{figure*}[t]
    \ContinuedFloat
    \centering

    \begin{tcolorbox}[
        title={Response from Our Method (Part 1) \hfill \textnormal{\scriptsize (Continued from previous page)}},
        colbacktitle=ourgreen!80!black, 
        coltitle=white, 
        colframe=ourgreen!80!black,
        colback=ourgreen!10!white,
        bottom=0mm 
    ]
        \scriptsize
       To solve this, I need to find the lower bound of the length of diagonal $BD$ of a rhombus inscribed in the hyperbola $\frac{x^2}{20} - \frac{y^2}{24} = 1$.
        
        \textbf{Step 1: Analyzing the Geometry}
        
        Let's think about the properties of the rhombus. Since the hyperbola is symmetric about the origin, and the rhombus is inscribed, its center must be the origin $(0,0)$.
        Let the vertices be $A, B, C, D$. The diagonals $AC$ and $BD$ must intersect at the origin.
        
        \textit{Let's try} to define the lines containing the diagonals.
        Crucially, the diagonals of a rhombus are perpendicular.
        If I let the slope of line $AC$ be $m$, then the line equation is $y = mx$.
        Consequently, the slope of line $BD$ must be $-1/m$, and its equation is $y = -\frac{1}{m}x$.
        
        \textbf{Step 2: Checking Constraints (Wait, do they strictly exist?)}
        
        I need to ensure that these lines actually intersect the hyperbola. Not all lines passing through the origin intersect a hyperbola.
        Let's substitute $y = kx$ into the hyperbola equation:
        \[ \frac{x^2}{20} - \frac{k^2x^2}{24} = 1 \implies x^2 \left( \frac{1}{20} - \frac{k^2}{24} \right) = 1 \]
        
        For $x$ to be a real number, we must have $x^2 > 0$. This implies:
        \[ \frac{1}{20} - \frac{k^2}{24} > 0 \implies \frac{1}{20} > \frac{k^2}{24} \implies k^2 < \frac{24}{20} = \frac{6}{5} \]
        
        \textit{Wait, let me double-check.} Does this apply to both diagonals?
        Yes, for the rhombus to exist, all four vertices must be on the hyperbola.
        So, the condition must hold for both the slope of $AC$ (which is $m$) and the slope of $BD$ (which is $-1/m$).
        
        Let's check the constraints for both:
        \begin{enumerate}
            \item For $AC$: $m^2 < \frac{6}{5}$
            \item For $BD$: $(-\frac{1}{m})^2 < \frac{6}{5} \implies \frac{1}{m^2} < \frac{6}{5} \implies m^2 > \frac{5}{6}$
        \end{enumerate}
        
        So, combining these, the range for $m^2$ is $\frac{5}{6} < m^2 < \frac{6}{5}$.
        This looks correct. If $m^2$ is outside this range, one of the diagonals won't hit the hyperbola.
        
        \textbf{Step 3: Formulating the Target Function}
        
        I need to minimize $BD^2$. Let's express the coordinates of point $B$.
        Point $B$ lies on the line $y = -\frac{1}{m}x$. Let's set $k_{bd} = -1/m$.
        From the intersection equation earlier:
        \[ x_B^2 = \frac{1}{\frac{1}{20} - \frac{k_{bd}^2}{24}} \]
        
        The squared length $BD^2$ is $(2 \cdot OB)^2 = 4(x_B^2 + y_B^2) = 4 x_B^2 (1 + k_{bd}^2)$.
        Let's try to simplify this expression.
        Substituting $k_{bd} = -1/m$, note that $k_{bd}^2 = 1/m^2$.
        Let $u = m^2$. Then the slope squared for BD is $1/u$.
        
        Wait, calculating with $1/u$ might be messy. Let's stick to the slope of $BD$ directly. Let $S = \text{slope of } BD$. Then $S^2 = 1/m^2$.
        From the inequality $\frac{5}{6} < m^2 < \frac{6}{5}$, we have:
        \[ \frac{5}{6} < \frac{1}{S^2} < \frac{6}{5} \implies \frac{5}{6} < S^2 < \frac{6}{5} \]
        Actually, the problem is symmetric. It doesn't matter which diagonal is which. Let's just say the slope squared of the diagonal of interest is $k^2$, where $k^2 \in (5/6, 6/5)$.
        
        Let's re-calculate $BD^2$ with slope $k$:
        \[ BD^2 = 4 x^2 (1+k^2) = \frac{4(1+k^2)}{\frac{1}{20} - \frac{k^2}{24}} = \frac{4(1+k^2)}{\frac{6-5k^2}{120}} = \frac{480(1+k^2)}{6-5k^2} \]
        \textbf{Step 4: Finding the Minimum (Trial and Computation)}
        
        Now I need to minimize $f(t) = \frac{480(1+t)}{6-5t}$ for $t \in (5/6, 6/5)$.
        Let's analyze the monotonicity.
        As $t$ increases, the numerator $(1+t)$ increases.
        The denominator $(6-5t)$ decreases (and stays positive since $t < 1.2$).
        Since the numerator is growing and the denominator is shrinking, the whole fraction must be strictly increasing.
        
        \textit{Let me verify this quickly.}
        If the function is increasing, the minimum value should be approached as $t \to 5/6$.
        Wait, the interval is open $(5/6, 6/5)$. So we strictly can't reach $5/6$.
        The question asks for the "greatest real number less than $BD^2$". This implies we are looking for the infimum.
        
        So I just need to plug in $t = 5/6$ to find the limit.
        Let's use Python to verify the calculation and ensure I haven't made a silly arithmetic error.
    \end{tcolorbox}
    

\end{figure*}

\begin{figure*}[t]
    \centering
    \begin{tcolorbox}[
        title={Response from Our Method (Part 2)},
        colbacktitle=ourgreen!80!black, 
        coltitle=white, 
        colframe=ourgreen!80!black,
        colback=ourgreen!10!white,
        top=0mm 
    ]
        \scriptsize
        
        
        

\begin{lstlisting}
import numpy as np

def bd_squared(t):
    # derived formula: 480(1+t) / (6-5t)
    return (480 * (1 + t)) / (6 - 5 * t)

# The range for slope squared is (5/6, 6/5)
lower_bound_t = 5/6

# Calculate the limit value
limit_val = bd_squared(lower_bound_t)
print(f"Limit at lower bound: {limit_val}")

# Let's also check a value slightly above 5/6 to confirm it's increasing
test_val = bd_squared(5/6 + 0.001)
print(f"Test value (slightly inside): {test_val}")
\end{lstlisting}
\begin{lstlisting}[backgroundcolor=\color{white}, frame=none]
Limit at lower bound: 480.0
Test value (slightly inside): 481.309...
\end{lstlisting}

        \textbf{Step 5: Final Conclusion}
        
        The code confirms the limit is 480 and the function is increasing.
        Since $BD^2$ is strictly increasing on the interval, $BD^2 > 480$ for all valid rhombi.
        Thus, the greatest real number less than $BD^2$ is 480.

        The greatest real number that is less than $BD^2$ for all such rhombi is \(\boxed{480}\).
    \end{tcolorbox}
\end{figure*}

\section{Dataset License}
Our training data is based on the MATH dataset~\citep{hendrycks2021measuring}, and we evaluate our model on AIME24~\citep{numina_math_datasets},
MATH-500~\citep{hendrycks2021measuring}, AMC~\citep{numina_math_datasets}, Minerva~\citep{lewkowycz2022solving}, and OlympiadBench~\citep{huang2024olympicarena}. We strictly adhere to the licenses associated with each dataset.

\section{LLMs Usage Statement}
We employed a Large Language Model (LLM) to assist exclusively in the editorial stage of manuscript preparation. Its role was limited to refining phrasing, correcting grammar, and enhancing clarity and readability across different sections. The LLM had no involvement in formulating research ideas, designing experiments, or conducting analyses. All scientific contributions and findings are entirely the work of the authors. The authors have ensured that the use of the LLM complies with ethical standards, avoiding plagiarism and scientific misconduct.

\end{document}